\documentclass{zlab}

\usepackage[T1]{fontenc}
\usepackage{tgpagella}
\usepackage{mathpazo}
\usepackage{inconsolata}
\usepackage{xspace}
\usepackage{graphicx,amsmath,amssymb,hyperref}
\usepackage[authoryear,round]{natbib}
\usepackage{xcolor}
\usepackage{booktabs}
\usepackage{colortbl}
\usepackage{makecell}
\usepackage{microtype}
\usepackage{multirow}
\usepackage{subcaption}
\usepackage[normalem]{ulem}
\usepackage{enumitem}
\usepackage{etoolbox}
\usepackage{adjustbox}
\usepackage{etoc}
\usepackage{tcolorbox}
\usepackage{tikz}
\usepackage[bottom]{footmisc}
\definecolor{Scol}{HTML}{1F4E9C}
\definecolor{Pcol}{HTML}{7B1FA2}
\definecolor{Rcol}{HTML}{B28CFF}
\definecolor{Good}{HTML}{1565C0}
\definecolor{Bad}{HTML}{C62828}
\definecolor{Neutral}{gray}{0.5}
\definecolor{tblstructured}{HTML}{6098FF}
\definecolor{tblsparse}{HTML}{B28CFF}
\definecolor{bestcellpurple}{HTML}{F3E5F5}
\definecolor{methodblue}{HTML}{EEF4FB}
\definecolor{diffbetter}{HTML}{DCEAFE}
\definecolor{diffworse}{HTML}{FCE4E4}
\definecolor{diffneutral}{HTML}{ECECEC}
\definecolor{brandpurple}{HTML}{7B1FA2}
\definecolor{promptbg}{HTML}{F9F9F9}
\definecolor{promptborder}{HTML}{7B1FA2}
\definecolor{citecol}{HTML}{5B9BD5}
\definecolor{linkblue}{HTML}{307DF0}
\hypersetup{colorlinks=true, citecolor=citecol, linkcolor=citecol, urlcolor=citecol}
\newcommand{\best}[1]{\textcolor{brandpurple}{\textbf{#1}}}
\newcommand{\method}[1]{\textbf{#1}}
\newcommand{\Sx}[1]{\textcolor{Scol}{#1}}
\newcommand{\Px}[1]{\textcolor{Pcol}{#1}}
\newcommand{\Rx}[1]{\textcolor{Rcol}{#1}}
\newcommand{\up}[1]{\cellcolor{bestcellpurple}\textcolor{Good}{+#1}}
\newcommand{\down}[1]{\cellcolor{bestcellpurple}\textcolor{Bad}{-#1}}
\newcommand{\same}[1]{\cellcolor{bestcellpurple}\textcolor{Neutral}{#1}}
\newtcolorbox{finding}{
    colback=promptbg,
    colframe=promptborder,
    arc=2pt,
    boxrule=0.8pt,
    left=8pt, right=8pt, top=5pt, bottom=5pt,
    before skip=6pt, after skip=6pt,
    fontupper=\normalsize,
}
\title{Small LLMs: Pruning vs. Training from Scratch}
\date{}

\newenvironment{abstractblock}{%
  {\centering\large\bfseries Abstract\par}
  \vspace{0.1em}
  \begin{list}{}{%
      \setlength{\leftmargin}{2em}
      \setlength{\rightmargin}{2em}
      \setlength{\topsep}{0pt}
      \setlength{\parsep}{0pt}
  }
  \item[]
}{%
  \end{list}
  \par\normalfont\vspace{0.5em}
}

\begin{document}
\fancypagestyle{firststyle}{\fancyhead[L]{}\fancyhead[C]{}\fancyhead[R]{}\fancyfoot[L]{}\fancyfoot[C]{}\fancyfoot[R]{}\renewcommand{\headrulewidth}{    1pt}\renewcommand{\footrulewidth}{0pt}\renewcommand{\headrule}{\color{gray}\hrule height\headrulewidth}}
\thispagestyle{firststyle}
\begin{center}
    \vspace*{2em}
    {\LARGE\bfseries Small LLMs: Pruning vs. Training from Scratch\par}
    \vspace{1.0em}
    {\large Yufeng Xu$^{1,2}$ \quad Taiming Lu$^{1}$ \quad Kunjun Li$^{1}$ \quad Jiachen Zhu$^{2}$ \\[0.4em]
    Mingjie Sun$^{3}$ \quad Zhuang Liu$^{1}$ \par}
    \vspace{0.1em}
    {\normalfont\fontsize{11}{15}\selectfont $^{1}$Princeton University \quad $^{2}$New York University \quad $^{3}$Carnegie Mellon University\par}
\end{center}
\let\oldthefootnote\thefootnote
\renewcommand{\thefootnote}{\textdagger}
\let\thefootnote\oldthefootnote

\vspace{1.5em}

\begin{figure*}[hbt!]
    \centering
    \begin{minipage}[t]{0.485\linewidth}
        \centering
        \includegraphics[height=1.6cm,width=\linewidth,keepaspectratio]{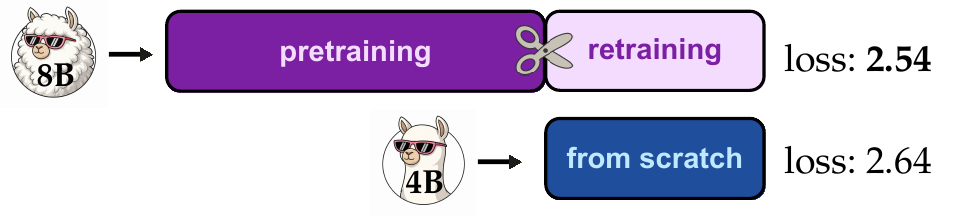}\\[0.5em]
        \includegraphics[height=4.0cm,width=\linewidth,keepaspectratio]{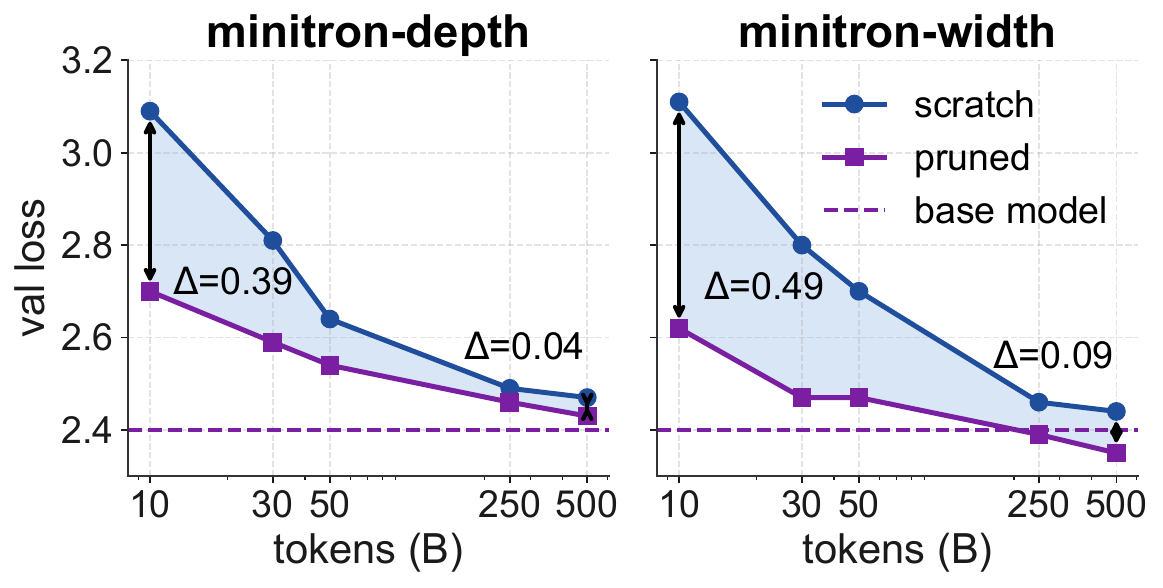}
    \end{minipage}\hfill
    \tikz[baseline=0pt]\draw[gray!55,dashed,line width=0.6pt] (0,1.4cm) -- (0,-3.8cm);\hfill
    \begin{minipage}[t]{0.485\linewidth}
        \centering
        \includegraphics[height=1.6cm,width=\linewidth,keepaspectratio]{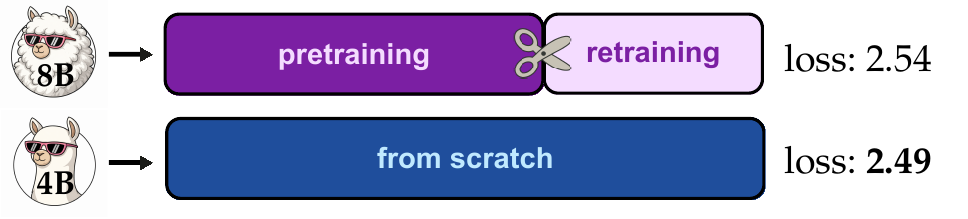}\\[0.5em]
        \includegraphics[height=4.0cm,width=\linewidth,keepaspectratio]{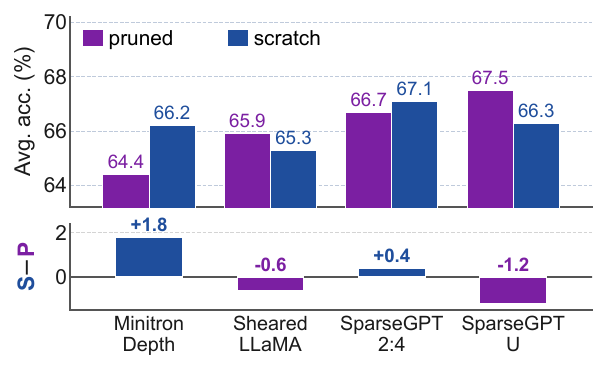}
    \end{minipage}
    \vspace{0em}
    \caption{
    \textbf{\Px{Initialization by pruning} provides a strong advantage over \Sx{random initialization}, but this advantage diminishes as training continues.}
    \textit{Left:} under the same training token budget, pruning initialization beats random initialization, although the advantage decreases with longer training.
    \textit{Right:} when the random initialization baseline is trained with the full token budget used by the entire pruning pipeline, it becomes competitive with pruning alternative.
    Overall, pruning is a powerful shortcut to strong small models, but its advantage diminishes as training scales.}
    \label{fig:teaser}
\end{figure*}

\vspace{0.7em}

\begin{abstractblock}
    Pruning promises a shortcut to strong small models.
    However, \citet{liu2019rethinkingvaluenetworkpruning} found that in CNNs, this shortcut often loses to training from scratch. LLMs are substantially different, with larger model and data scales and different architectures, so the same conclusions may not hold.
    We examine this by pruning Llama-3.1-8B at pruning ratios of 0.5--0.8 with six methods spanning depth, width, and sparse granularities, under two token-matched settings.
    \textbf{(1)} 
    With the same training token budget, pruned initialization consistently outperforms random initialization. This shows that the parent model provides a strong starting point, although the advantage narrows as the training token budget grows and the pruning ratio rises, nearly vanishing at the highest ratio we study.
    \textbf{(2)} 
    When training from scratch is instead given the full token budget consumed by the whole pipeline, pruning at finer granularities still retains an advantage, while coarser structured pruning can be matched or surpassed. This suggests that the parent model transfers knowledge that additional tokens alone cannot fully recover, but only at fine granularity.
    This forms a clear recommendation: with a strong parent and a limited token budget, pruning is more effective; when tokens are plentiful, training from scratch is competitive at coarse granularity, so a parent is not always necessary.
    Our code is available at \href{https://github.com/zlab-princeton/pruning-vs-scratch}{\textcolor[HTML]{307DF0}{github.com/zlab-princeton/pruning-vs-scratch}}.
\end{abstractblock}

\newpage

\section{Introduction}

Remove more than half of a trained language model's parameters, add a lightweight retraining pass, and the resulting small model still matches most of the original's performance~\citep{ma2023llmpruner, xia2024shearedllama, sun2024wanda, sreenivas2024minitron2}. This is the promise of network pruning~\citep{lecun1989OBD, babak1993obs}: a shortcut to strong small language models that bypasses the rising cost of pretraining Large Language Models (LLMs) from scratch on trillions of tokens~\citep{grattafiori2024llama3, qwen3, deepseekv3, kaplan2020scalinglaw, hoffmann2022computeoptimalllm, henighan2020scalinglaw, touvron2023llama, touvron2023llama2}.

Multi-billion-parameter open-weight LLMs are now widely available, with releases such as Llama~\citep{grattafiori2024llama3, meta2025llama3.1}, Qwen~\citep{qwen3}, Gemma~\citep{gemma3}, and OLMo~\citep{olmo3} putting pretrained checkpoints in public hands. At the same time, demand for smaller models is rising, driven by deployment constraints, inference efficiency, and accessibility~\citep{hagele2024scalinglawscomputeoptimaltraining, hu2024minicpm, nag2024efficientcontinualpretrainingllms}. This is exactly the gap pruning promises to fill. This raises a practical question: to obtain a capable small model, should we train it from scratch, or carve it out of an existing large one via pruning? Pruning is appealing because it may transfer knowledge from the parent, giving a stronger starting point than random initialization. But the inheritance is not free: a large parent must be pretrained before it can be pruned, and whether pruning's advantage survives a token-matched comparison remains unclear.

While pruning is typically viewed as a compression technique, we instead treat it as an \emph{initialization strategy}: the pruned weights are simply a starting point for training the target small model. This reframes the natural baseline: rather than comparing the pruned model against its parent, we compare it against a model of the same size trained from a random initialization on the same data stream. The question becomes: under an equal training token budget, does \textit{initializing by pruning a larger model} beat \textit{random initializing}?

\citet{liu2019rethinkingvaluenetworkpruning} first examined this question for convolutional networks on image-classification benchmarks. For \emph{structured} pruning, they found that a target architecture trained from \emph{scratch} matches or surpasses the same architecture pruned and fine-tuned from a larger model, implying that pruning's value is the \emph{architecture} it discovers rather than the weights it inherits, and recasting structured pruning as a form of neural architecture search. For \emph{unstructured} pruning, however, the inherited weights were harder to replace, with training from scratch falling short of the pruned-and-fine-tuned model at larger scale. This suggests that pruning's benefit shifts with granularity, from the architecture alone at the coarse end toward the specific inherited weights at the fine end. Whether this carries over to LLMs is unclear, given a different regime: larger data and model scales, different architectures, and higher demands for compact deployable models.

We therefore ask a sharper question: are pruning's gains merely a head start that more training can buy, or knowledge that extra data cannot recover? To find out, we compare pruned against random initialization under two token-matched settings. \textbf{(i)~Equal training token budget}: both target models are trained under the same training token budget, isolating the effect of initialization. \textbf{(ii)~Equal total token budget}: training from scratch is instead given the \emph{entire} token budget of the pruning pipeline (the parent's pretraining tokens plus the retraining tokens), testing whether more tokens alone can close the gap. Varying pruning ratio and granularity across both, we draw two key observations, both illustrated in Figure~\ref{fig:teaser}:

  \begin{enumerate}
      \item[(1)] Under an equal training token budget, training from pruned initialization consistently outperforms training from random initialization, though this advantage diminishes as pruning ratio increases.
      \item[(2)] Even when training from scratch uses the full token budget (number of training iterations) of the pruning pipeline, sparse pruning still outperforms; for structured pruning, however, training from scratch can close the gap.
  \end{enumerate}

    Together, these results recast pruning as a token-efficient initialization strategy: with a strong parent model in hand and a limited training token budget, pruning beats training from scratch, especially at fine granularity, transferring knowledge that additional tokens alone cannot recover.

\begin{figure}[!t]
    \centering
    \begin{minipage}[t]{0.31\linewidth}
        \centering
        \vspace{0pt}
        {\small\textbf{(a) Pruning granularity}\par}
        \vspace{3pt}
        \includegraphics[width=0.92\linewidth]{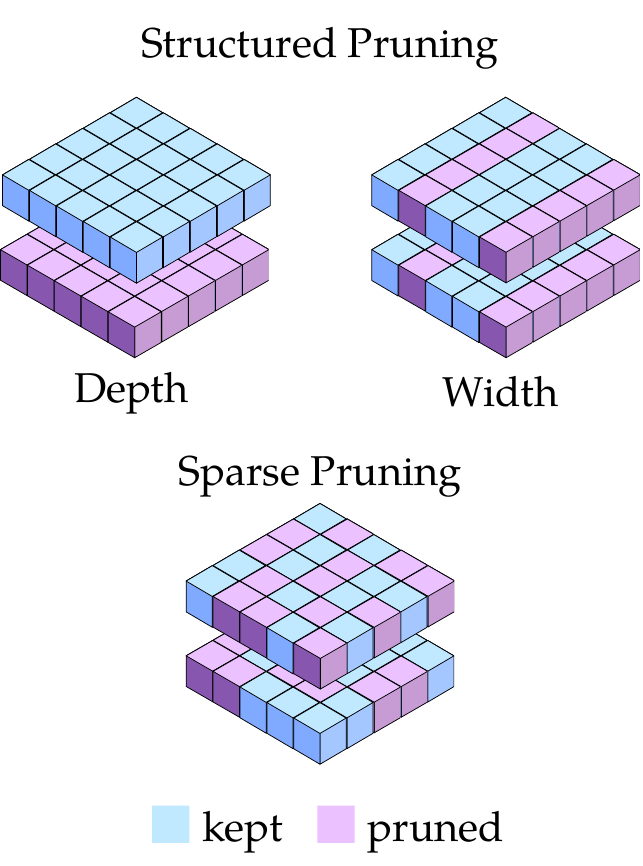}
    \end{minipage}
    \hfill
    \begin{minipage}[t]{0.65\linewidth}
        \centering
        \vspace{0pt}
        {\small\textbf{(b) Methods evaluated}\par}
        \vspace{10pt}
        {\fontsize{8.4}{11}\selectfont
        \renewcommand{\arraystretch}{1.18}
        \setlength{\tabcolsep}{3pt}
        \begin{tabular}{lll}
            \toprule
            method & target modules & criterion \\
            \midrule
            \makecell[l]{Minitron\\depth} & \makecell[l]{layers} & \makecell[l]{influence on\\validation loss} \\
            \midrule
            \makecell[l]{Minitron\\width} & \makecell[l]{hidden channels,\\mlp channels} & \makecell[l]{block output\\activation norm} \\
            \midrule
            \makecell[l]{FLAP} & \makecell[l]{hidden channels,\\mlp channels} & \makecell[l]{input feature variance\\$\times$ weight column norm} \\
            \midrule
            \makecell[l]{Sheared\\LLaMA} & \makecell[l]{layers, hidden channels,\\mlp channels, attn heads} & \makecell[l]{learned pruning masks\\(Lagrangian optim.)} \\
            \midrule
            \makecell[l]{Wanda} & \makecell[l]{individual weights} & \makecell[l]{weight magnitude\\$\times$ input feature norm} \\
            \midrule
            \makecell[l]{SparseGPT} & \makecell[l]{individual weights} & \makecell[l]{layerwise\\reconstruction error} \\
            \bottomrule
        \end{tabular}
        }
    \end{minipage}
    \vspace{-1pt}
    \caption{\textbf{Pruning granularity and method overview.} (a) Illustration of the pruning granularities we study: depth, width, and sparse pruning. (b) Overview of the six pruning methods we evaluate, including their target modules and criteria.}
    \label{fig:prune-methods}
    \vspace{-.2cm}
\end{figure}

\section{Background}
\label{sec:background}

Network pruning~\citep{lecun1989OBD} identifies and removes unimportant weights or modules in neural networks to improve efficiency~\citep{frankle2019LTH}. Methods fall into two families (Figure~\ref{fig:prune-methods}~(a)): \textit{sparse pruning}, which zeroes out individual weights while preserving model shape, and \textit{structured pruning}, which removes entire architectural components such as layers, attention heads, or feedforward channels.

\paragraph{Sparse pruning.}
Sparse methods mask individual weights to zero. \textit{Semi-structured} (n:m) pruning~\citep{zhou2021nmpruning} keeps exactly $n$ of every $m$ weights and can exploit hardware support~\citep{mishra2021acceleratingsparse, hubara2021acceleratedsparsetraining}; \textit{unstructured} pruning places no constraint on which weights are removed. SparseGPT~\citep{frantar2023sparsegpt} frames weight selection as a layerwise reconstruction problem inspired by Optimal Brain Surgeon~\citep{babak1993obs}. Wanda~\citep{sun2024wanda} approximates the same objective more cheaply by scoring each weight as its magnitude times the corresponding input feature norm, with follow-up work~\citep{yang2025wanda++} further refining the activation-based scoring. Both methods support n:m and unstructured modes and require only a lightweight weight update after pruning.

\paragraph{Structured pruning.}
Structured methods, with roots in classical CNN-era channel and filter pruning~\citep{li2017l1normpruning, liu2017networkslimming, luo2017thinet, huang2018sparsestructureselection}, remove whole modules, yielding smaller dense architectures. \textit{Depth pruning} removes entire layers: ShortGPT~\citep{men2024shortgpt}, Shortened Llama~\citep{kim2024shortenedllama}, and SLEB~\citep{song2024sleb} estimate layer importance via activation similarity, Taylor approximations, and block-skip perplexity respectively; Minitron-depth~\citep{sreenivas2024minitron2} searches for the consecutive layer group whose removal least hurts validation loss. \textit{Width pruning} removes hidden or feedforward channels: Minitron-width~\citep{sreenivas2024minitron2} ranks channels by output activation norm, while FLAP~\citep{an2023flap} uses input feature variance weighted by column norm. Sheared LLaMA~\citep{xia2024shearedllama} jointly prunes both depth and width by learning pruning masks via Lagrangian optimization subject to a fixed target model size constraint specified in advance.

\paragraph{Pruning ratio and granularity.}
The \textit{pruning ratio} is the fraction of parameters removed. \textit{Granularity} refers to the smallest unit treated as a single pruning decision, ranging from whole layers (coarsest) through channels and attention heads down to individual weights (finest). Under the same ratio, finer granularity preserves more of the base model's performance but yields less hardware speedup and reduced memory savings at inference time~\citep{an2023flap, sreenivas2024minitron2}.

\section{Methodology}
\label{sec:methodology}

We study pruning under constrained token budgets through controlled comparisons that isolate (i) the value of pruning as an initialization strategy and (ii) whether additional training from scratch can close the gap to the pruning pipeline. Section~\ref{subsec:training-setup} describes the experimental design, base model, data pipeline, optimizer, and evaluation protocol. Section~\ref{subsec:pruning-methods} formalizes the six pruning methods we evaluate.

\subsection{Training Setup}
\label{subsec:training-setup}

\paragraph{Experimental design.} Throughout, we use three notations: \Sx{S}$N$ denotes training the target architecture from \Sx{S}cratch with random initialization for $N$B tokens; \Px{P200}-\Rx{R}$N$ denotes the pruning pipeline, which \Px{P}retrains a larger model for \Px{200}B tokens, prunes it at a target ratio, then \Rx{R}etrains the pruned model for $N$B tokens; and \emph{Meta}-\Rx{R}$N$ denotes the same pipeline but starting from Meta's released Llama-3.1-8B checkpoint instead of our 200B-pretrained one, retrained for $N$B tokens (used as a reference point in Figure~\ref{fig:minitron-token-scaling}). Both token-matched experiments compare the pruning pipeline against training from scratch, differing only in how much data we allow the scratch baseline to see:
\begin{itemize}
    \item[(1)] \textbf{Equal training token budget} (\Sx{S50} vs.\ \Px{P200}-\Rx{R50}): the scratch baseline trains on the same 50B-token data stream used to retrain the pruned model, under the same target architecture and optimizer schedule, isolating the effect of initialization alone. We further sweep the pruning ratio to track how this initialization advantage evolves as compression becomes more aggressive.
    \item[(2)] \textbf{Equal total token budget} (\Sx{S250} vs.\ \Px{P200}-\Rx{R50}): the scratch baseline trains on all 250B tokens consumed by the pipeline (pretraining plus retraining), testing whether additional tokens alone can close the gap.
\end{itemize}

\paragraph{Model and dataset.} We adopt Llama-3.1-8B~\citep{grattafiori2024llama3} as the base model for our experiments, which is the most common~\citep{frantar2023sparsegpt, sun2024wanda, sreenivas2024minitron2} choice in prior LLM pruning works. For training data, we adopt DCLM-Baseline-1.0~\citep{li2024datacomplm}, a large-scale curated training corpus comparable to other contemporary web-scale corpora such as RefinedWeb~\citep{penedo2023refinedweb}, Dolma~\citep{soldaini2024dolma}, and RedPajama~\citep{weber2024redpajama}.
The dataset is tokenized with the Llama-3.1 tokenizer and divided into chunks of 8192 tokens.

\paragraph{Data pipeline.} The 200B pretraining tokens and the 50B retraining tokens are strictly non-overlapping subsets of the DCLM corpus. For the \Sx{S250} comparison, the 250B tokens are exactly the union of these two disjoint sets, ensuring that \Sx{S250} and \Px{P200}-\Rx{R50} see exactly the same data in total. For the \Sx{S50} comparison, \Sx{S50} uses the same 50B tokens as the retraining phase of \Px{P200}-\Rx{R50}. This guarantees that all comparisons are token-fair and that differences in performance reflect initialization strategy, not data composition.

\paragraph{Training recipe.} We follow the default recipe in Lingua~\citep{meta2024lingua}, using AdamW~\citep{loshchilov2019adamw} with cosine learning rate decay and a 5\% warmup. For each experiment we sweep the max learning rate over \texttt{\{1e-5,3e-5,1e-4,3e-4,1e-3\}} and select the one yielding lowest training loss in a short prefix run. This is important for retraining after pruning, where the learning rate must be chosen carefully to prevent catastrophic forgetting~\citep{parmar2024reusedontretrainrecipe, gupta2023howtorewarmyourmodel}. The full set of training hyperparameters is reported in Appendix~\ref{appendix:training-recipe}.

\paragraph{Evaluation.} \label{sec:exp-eval} We evaluate on two dimensions: (i) linguistic perplexity across four general-domain corpora (C4, WikiText-103, WikiText-2, CNN Dailymail) and (ii) zero-shot downstream accuracy across eight benchmarks spanning commonsense QA (WinoGrande, HellaSwag, PIQA), scientific QA (ARC-Challenge, ARC-Easy, SciQ, OpenBookQA), and logical reasoning (BoolQ). Full evaluation protocols, benchmark citations, and random-chance baselines are provided in Appendix~\ref{appendix:evaluation}.

\subsection{Pruning Methods}
\label{subsec:pruning-methods}

We evaluate six representative LLM pruning methods spanning two families, with their target modules and scoring criteria summarized in Figure~\ref{fig:prune-methods}~(b) (and described in Section~\ref{sec:background}).
\textit{Structured} methods remove entire model components while keeping the weight matrix dense (\method{Minitron-depth}~\citep{muralidharan2024minitron1, sreenivas2024minitron2}, \method{Minitron-width}~\citep{muralidharan2024minitron1, sreenivas2024minitron2}, \method{FLAP}~\citep{an2023flap}, and \method{Sheared LLaMA}~\citep{xia2024shearedllama}); \textit{sparse} methods retain the full macro-architecture but zero-mask individual weights (\method{Wanda}~\citep{sun2024wanda} and \method{SparseGPT}~\citep{frantar2023sparsegpt}).
For brevity, we refer to Minitron-depth as \method{Minitron-D}, Minitron-width as \method{Minitron-W}, and the unstructured variants of Wanda and SparseGPT as \method{Wanda-U} and \method{SparseGPT-U}, respectively.
Formal mathematical descriptions of all six methods are provided in Appendix~\ref{appendix:pruning-method-details}.

\paragraph{Implementation.}
For Minitron-depth and Minitron-width, we developed a simplified re-implementation and verified its correctness against the original results.
FLAP and Sheared LLaMA were originally designed for models with multi-head attention (MHA); we adapted both methods to support grouped-query attention (GQA)~\citep{ainslie2023gqa} as used in Llama-3.1, and verified that they achieve reasonable pruning performance on this architecture.
For Wanda and SparseGPT, we used the authors' original implementations without modification.
When a target architecture was not specified in the original paper, we performed a lightweight architecture search over candidate dimension configurations (see Appendix~\ref{appendix:pruning-details}).
Full per-method calibration sets and hyperparameters are reported in Appendix~\ref{appendix:pruning-details}.

\begin{table}[!t]
    \centering
    {\fontsize{9}{11}\selectfont
    \begin{tabular}{lccccccc}
        \specialrule{1.2pt}{3pt}{5pt}
        method & ratio & num layers & attention heads & hidden size & FFN size & sparsity & parameters \\
        \specialrule{1.2pt}{3pt}{0pt}
        base & 0.0\% & 32 & 32 & 4096 & 14336 & 0.0\% & 8.0B \\
        \noalign{\vspace{-3pt}}\midrule
        Minitron-D &50.0\% & 16 & 32 & 4096 & 14336 & 0.0\% & 4.5B \\
        Minitron-D &62.5\% & 8 & 32 & 4096 & 14336 & 0.0\% & 2.8B \\
        Minitron-D &75.0\% & 4 & 32 & 4096 & 14336 & 0.0\% & 1.9B \\
        Minitron-D &81.3\% & 2 & 32 & 4096 & 14336 & 0.0\% & 1.5B \\
        \midrule
        Minitron-W &50.0\% & 32 & 32 & 3072 & 9216 & 0.0\% & 4.5B \\
        Minitron-W &62.5\% & 32 & 32 & 2432 & 6144 & 0.0\% & 3.1B \\
        Minitron-W &75.0\% & 32 & 32 & 1792 & 6016 & 0.0\% & 2.0B \\
        Minitron-W &81.3\% & 32 & 32 & 1536 & 4736 & 0.0\% & 1.5B \\
        \midrule
        FLAP & 50.0\% & 32 & 20 & 4096 & 6656 & 0.0\% & 4.5B \\
        \midrule
        Sheared LLaMA & 50.0\% & 32 & 32 & 3456 & 7552 & 0.0\% & 4.6B\\
        \midrule
        Wanda & 50.0\% & 32 & 32 & 4096 & 14336 & 50.0\% & 4.5B \\
        \midrule
        SparseGPT & 50.0\% & 32 & 32 & 4096 & 14336 & 50.0\% & 4.5B\\
        \bottomrule
    \end{tabular}
    }
    \vspace{2pt}
    \caption{Architectures of the pruned models obtained with different pruning methods and pruning ratios. The base model is Llama-3.1-8B (8.0B parameters). For sparse methods the listed parameter count reflects non-zero weights only.}
    \label{tab:model-configs}
    \vspace{-.3cm}
\end{table}

\paragraph{Pruning configurations.}
The six methods span five granularities: depth, width, depth-and-width, 2:4 sparse, and unstructured sparse. We center our comparison at a 50\% pruning ratio because it is the canonical setting adopted by prior pruning work---Minitron~\citep{sreenivas2024minitron2}, SparseGPT~\citep{frantar2023sparsegpt}, and Wanda~\citep{sun2024wanda} all report their primary results at 50\%---so we prune every method to this same ratio and search for the best configuration within each method's design space. Concretely, we evaluate Minitron-depth at $\{50\%, 62.5\%, 75\%, 81.3\%\}$ pruning ratios, Minitron-width at $\{50\%, 62.5\%, 75\%\}$, FLAP and Sheared LLaMA at $50\%$; SparseGPT and Wanda are applied at 2:4 and 50\% unstructured sparsity. Resulting architectures are listed in Table~\ref{tab:model-configs}. Calibration sets and per-method configurations are in Appendix~\ref{appendix:pruning-details}.

\section{Pruning vs. Training from Scratch}

We evaluate pruning under the equal-training-token-budget setting introduced in Section~\ref{sec:methodology}, comparing P200-R50 against S50. We also include S250 to see how training from scratch with the full pipeline token budget compares to pruning. Beyond the 50\% pruning ratio, we additionally vary the pruning ratio for Minitron-D and Minitron-W to understand how the advantage of pruned initialization changes with more aggressive compression. Table~\ref{table:main-50pct-summary} summarizes the comparison of P200-R50 against S50 and S250 across all six pruning methods (with both 2:4 and unstructured-sparsity variants for Wanda and SparseGPT) at 50\% pruning ratio, with the full per-benchmark breakdown deferred to Appendix~\ref{appendix:equal-total-data} (Table~\ref{table:pruning-vs-scratch-50pct}). Among the six methods, \method{Wanda-U} (the finest sparse granularity) yields the most accurate pruned model (68.1\% average accuracy) and \method{FLAP} the lowest perplexity. The accuracy ranking broadly tracks pruning granularity, though the most accurate (sparse) methods bring little inference speedup on commodity hardware---a trade-off we examine in Section~\ref{subsec:granularity}. Table~\ref{table:pruning-ratio-ablations} reports results across pruning ratios for Minitron-D and Minitron-W. We highlight two key observations.

\begin{table}[t]
\centering
\small
\setlength{\tabcolsep}{5pt}
\begin{tabular}{l ccc ccc}
\toprule
& \multicolumn{3}{c}{WikiText-2 ppl $\downarrow$} & \multicolumn{3}{c}{Avg accuracy (\%) $\uparrow$} \\
\cmidrule(lr){2-4}\cmidrule(lr){5-7}
method & \Sx{S50} & \Px{P200-R50} & \Sx{S250} & \Sx{S50} & \Px{P200-R50} & \Sx{S250} \\
\midrule
Minitron-D    & 10.77 & 9.41 & 9.01 & 60.7 & 64.4 & 66.2 \\
Minitron-W    & 11.17 & 9.01 & 8.69 & 59.9 & 66.3 & 67.5 \\
FLAP          & 11.34 & \textbf{7.97} & 8.82 & 57.7 & 66.5 & 65.1 \\
Sheared LLaMA & 11.28 & 8.63 & 8.85 & 58.4 & 65.9 & 65.3 \\
\midrule
Wanda-2:4     & 10.55 & 9.15 & 8.60 & 61.5 & 66.7 & 66.2 \\
Wanda-U       & 10.66 & 8.37 & 8.57 & 61.6 & \textbf{68.1} & 67.2 \\
SparseGPT-2:4 & 10.54 & 8.90 & 8.65 & 62.1 & 66.7 & 67.1 \\
SparseGPT-U   & 10.47 & 8.44 & 8.59 & 61.9 & 67.5 & 66.3 \\
\bottomrule
\end{tabular}
\caption{\textbf{Pruning--retraining vs.\ training from scratch at 50\% pruning ratio (summary).} WikiText-2 perplexity ($\downarrow$) and average downstream accuracy ($\uparrow$, over the eight benchmarks) for \Px{P200-R50} (pretrain 200B, prune, retrain 50B), \Sx{S50} (scratch for 50B; equal training token budget), and \Sx{S250} (scratch for 250B; equal total token budget). \textbf{Bold} marks the best pruned-model (\Px{P200-R50}) result per metric. \textit{Pruning beats equal-training-token-budget scratch (\Sx{S50}) for every method; under an equal total token budget (\Sx{S250}), only the finer (sparse) granularities keep their edge, while coarser structured pruning is matched.} Full per-benchmark results are in Table~\ref{table:pruning-vs-scratch-50pct}.}
\label{table:main-50pct-summary}
\end{table}

\begin{figure*}[!t]
    \centering
    \begin{subfigure}[t]{0.49\linewidth}
        \centering
        \includegraphics[width=\linewidth]{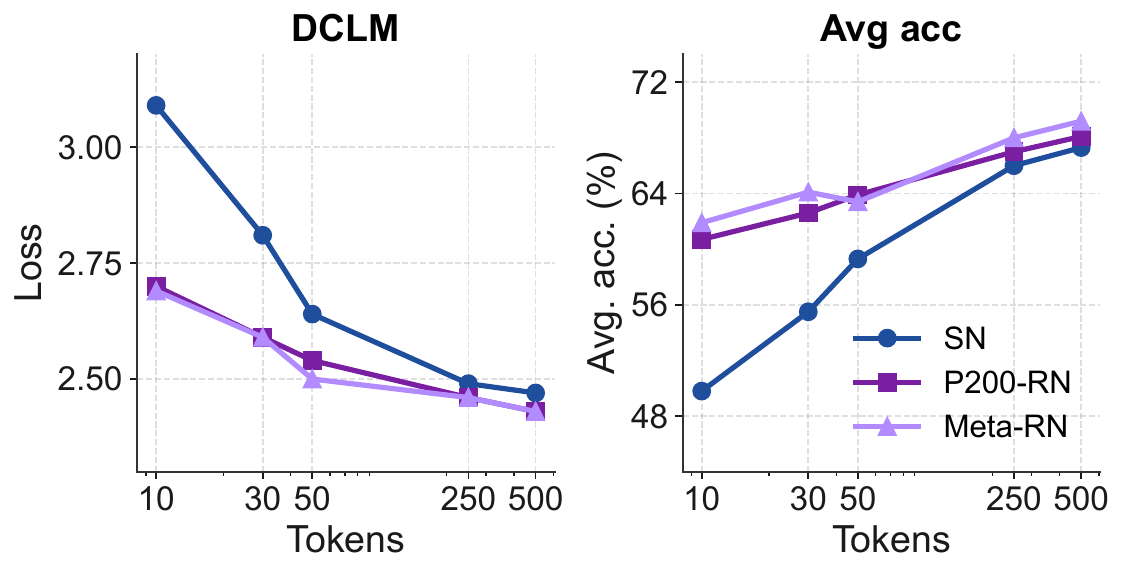}
        \caption{Minitron-D pruning.}
        \label{fig:minitron-depth-results}
    \end{subfigure}\hfill
    \begin{subfigure}[t]{0.49\linewidth}
        \centering
        \includegraphics[width=\linewidth]{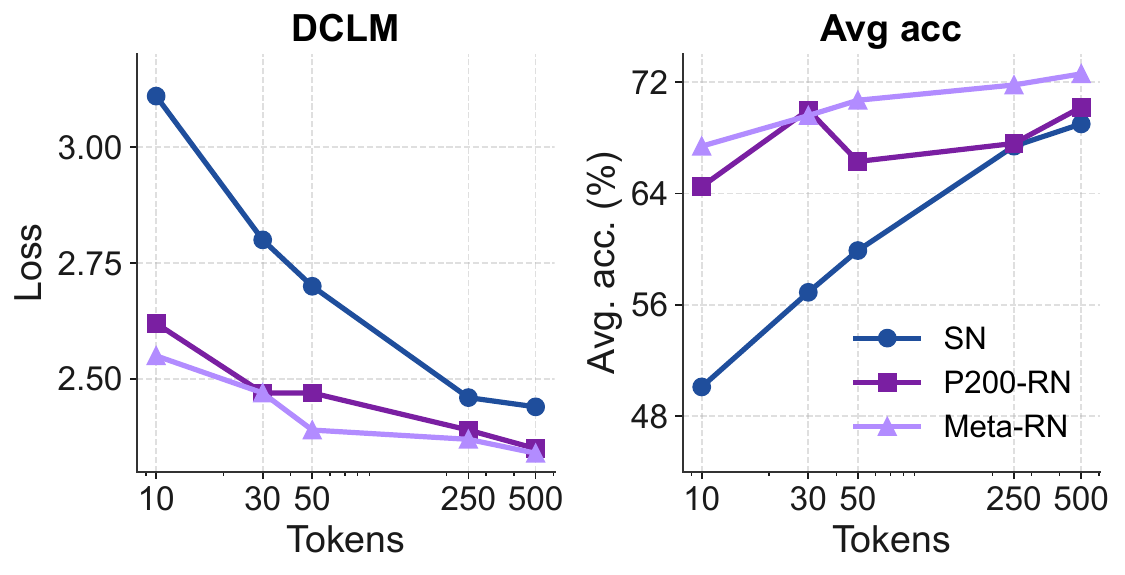}
        \caption{Minitron-W pruning.}
        \label{fig:minitron-width-results}
    \end{subfigure}
    \\[4pt]
    \begin{subfigure}[t]{0.49\linewidth}
        \centering
        \includegraphics[width=\linewidth]{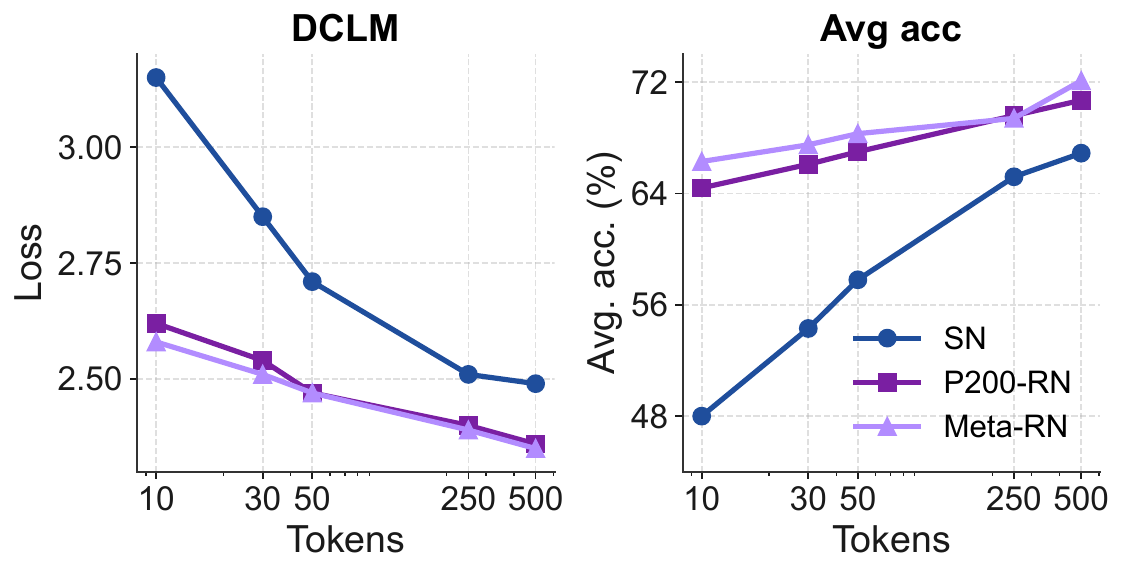}
        \caption{FLAP pruning.}
        \label{fig:flap-results}
    \end{subfigure}\hfill
    \begin{subfigure}[t]{0.49\linewidth}
        \centering
        \includegraphics[width=\linewidth]{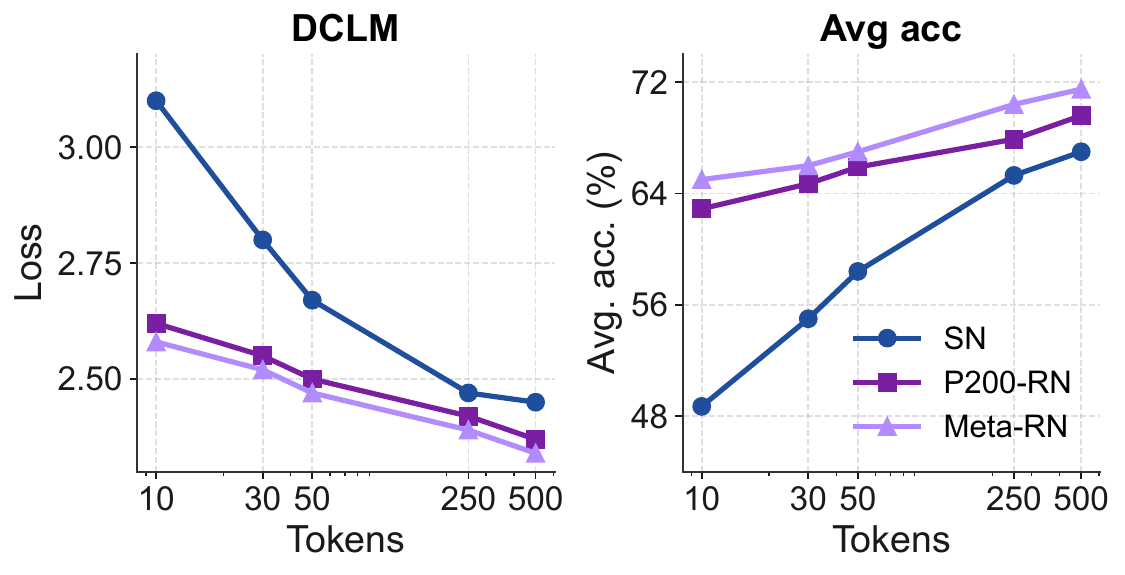}
        \caption{Sheared LLaMA pruning.}
        \label{fig:sheared-results}
    \end{subfigure}
    \caption{\textbf{Four structured pruning methods across retraining token budgets (Llama-3.1-8B $\to$ 4B, 50\% pruning).} Each panel shows DCLM validation loss ($\downarrow$) and average downstream accuracy ($\uparrow$; averaged over WinoGrande, ARC-C, ARC-E, HellaSwag, PIQA, SciQ, BoolQ, OBQA) as a function of retraining tokens $N \in \{10, 30, 50, 250, 500\}$B under three initialization strategies: \textit{S$N$} (train from scratch for $N$B tokens), \textit{P200-R$N$} (pretrain 200B tokens, prune, retrain for $N$B tokens), and \textit{Meta-R$N$} (prune from Meta's released Llama-3.1-8B, retrain for $N$B tokens). \textit{Pruned initialization (P200-R$N$) starts well ahead of scratch (S$N$) and keeps its lead as retraining grows, though the gap narrows with more tokens.} Full results are in Appendix~\ref{appendix:token-scaling}.}
    \label{fig:minitron-token-scaling}
\end{figure*}

\subsection{Equal Training Token Budget}

\begin{finding}
\textbf{Finding 1.} Under an equal training token budget, pruning initialization consistently beats random initialization, but the advantage shrinks as the pruning ratio grows and vanishes near 81\% sparsity.
\end{finding}

\paragraph{Pruning initialization is consistently better than random initialization under 50\% pruning ratio.} \label{subsec:prune-better-random}
The experiment results under 50\% pruning ratio are summarized in Table~\ref{table:main-50pct-summary} (full per-benchmark breakdown in Table~\ref{table:pruning-vs-scratch-50pct}); Figure~\ref{fig:minitron-token-scaling} plots DCLM loss and average accuracy across retraining token budgets for all four structured pruning methods (with full per-metric results in Tables~\ref{table:minitron-depth-token-scaling},~\ref{table:minitron-width-token-scaling},~\ref{table:flap-token-scaling}, and~\ref{table:sheared-token-scaling} in Appendix~\ref{appendix:token-scaling}), and the $N{=}50$ slice matches the equal-training-token-budget comparison discussed here. On structured pruning methods, Minitron-D combined with P200-R50 consistently outperforms S50 on all 8 benchmarks, with significant 4.5\% and 6.1\% improvement on ARC Challenge and Hellaswag (Figure~\ref{fig:minitron-depth-results}). On Minitron-W, the improvement of P200-R50 over S50 is even more significant, with 8.2\% improvement on ARC Challenge and 10.2\% improvement on Hellaswag (Figure~\ref{fig:minitron-width-results}), which is consistent with the observation that Minitron-W preserves more of the base model's capabilities than Minitron-D~\citep{sreenivas2024minitron2}. FLAP and Sheared LLaMA exhibit the same qualitative trend across token budgets (Figures~\ref{fig:flap-results} and~\ref{fig:sheared-results}).

Sparse pruning follows the same pattern, with the finer unstructured granularity gaining more than 2:4: for Wanda, P200-R50 improves WikiText-2 perplexity over S50 by 1.40 (2:4) versus 2.30 (unstructured), and for SparseGPT the ARC-Challenge gain grows from 2.0\% (2:4) to 8.1\% (unstructured). This unstructured-over-2:4 ordering holds across nearly all benchmarks (Table~\ref{table:main-50pct-summary}).

\paragraph{The advantage of pruning initialization diminishes as pruning ratio increases.}
Table~\ref{table:pruning-ratio-ablations} (Appendix~\ref{appendix:pruning-ratio}) presents results across different pruning ratios for Minitron-D and Minitron-W. At 50\% pruning ratio, P200-R50 shows a clear advantage over S50, with 1.31 lower perplexity on WikiText-103. However, this advantage shrinks as pruning becomes more aggressive: at 62.5\% and 75\% pruning ratio, the gap narrows substantially. At 81.3\% pruning ratio, P200-R50 performs on par with or even slightly worse than S50 on some benchmarks, suggesting that pruning initialization loses its benefit under extreme compression.

\begin{figure*}[t!]
    \centering
    \begin{subfigure}[t]{0.49\linewidth}
        \centering
        \includegraphics[width=\linewidth]{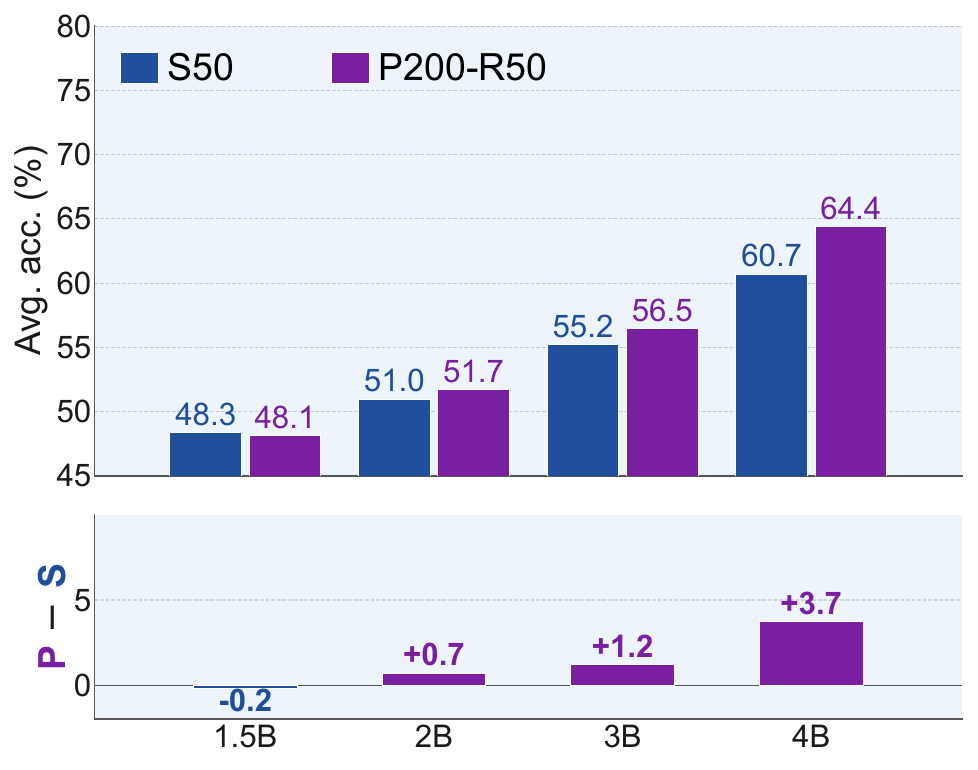}
        \caption{Depth pruning: S50 vs.\ P200-R50 across model sizes.}
        \label{fig:depth-s50-l200s50}
    \end{subfigure}\hfill
    \begin{subfigure}[t]{0.49\linewidth}
        \centering
        \includegraphics[width=\linewidth]{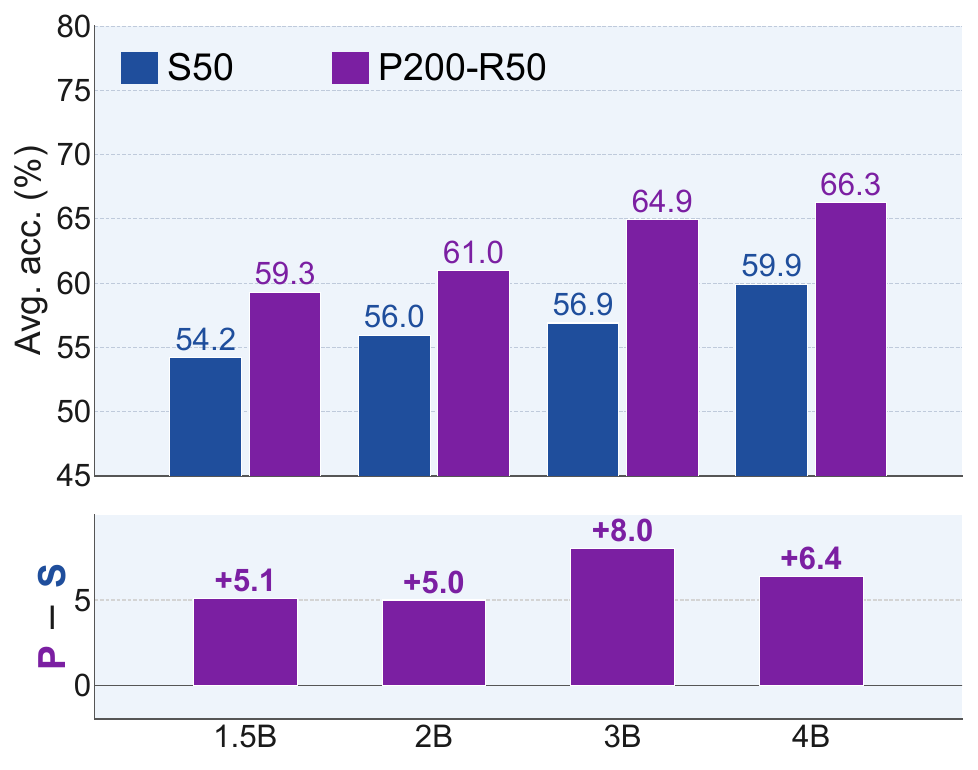}
        \caption{Width pruning: S50 vs.\ P200-R50 across model sizes.}
        \label{fig:width-s50-l200s50}
    \end{subfigure}
    \vspace{-.2cm}
    \caption{\textbf{S50 vs.\ P200-R50 across model sizes for depth and width pruning.} Top panels show average accuracy (WinoGrande, ARC-C, ARC-E, HellaSwag, PIQA, SciQ, BoolQ, OBQA); bottom panels show the gain of P200-R50 over S50 per model size. Dark blue bars denote scratch-trained models (S50); purple bars denote pruned-then-retrained models (P200-R50) at the same target parameter count. \textit{The pruning advantage shrinks as the target model gets smaller (i.e., as the pruning ratio grows), nearly closing at the smallest sizes.}}
    \label{fig:size-scaling}
\end{figure*}

On perplexity benchmarks, the gap between P200-R50 and S50 closes as pruning ratio increases from 50\% to 81.3\%, with the curves eventually intersecting. Similar patterns are observed on accuracy benchmarks, though with some exceptions such as Winogrande. Figure~\ref{fig:size-scaling} visualizes the same trend across pruned model sizes for both depth pruning (Figure~\ref{fig:depth-s50-l200s50}) and width pruning (Figure~\ref{fig:width-s50-l200s50}): the advantage of P200-R50 over S50 shrinks as the target model becomes smaller (i.e., as the pruning ratio grows).

\subsection{Can Extended Training Catch Up?}

\begin{finding}
\textbf{Finding 2.} Given the pipeline's full token budget, training from scratch catches up to pruning at coarser granularities, but pruning at finer granularities retains its advantage.
\end{finding}

\begin{figure*}[t!]
    \centering
    \includegraphics[width=0.95\linewidth]{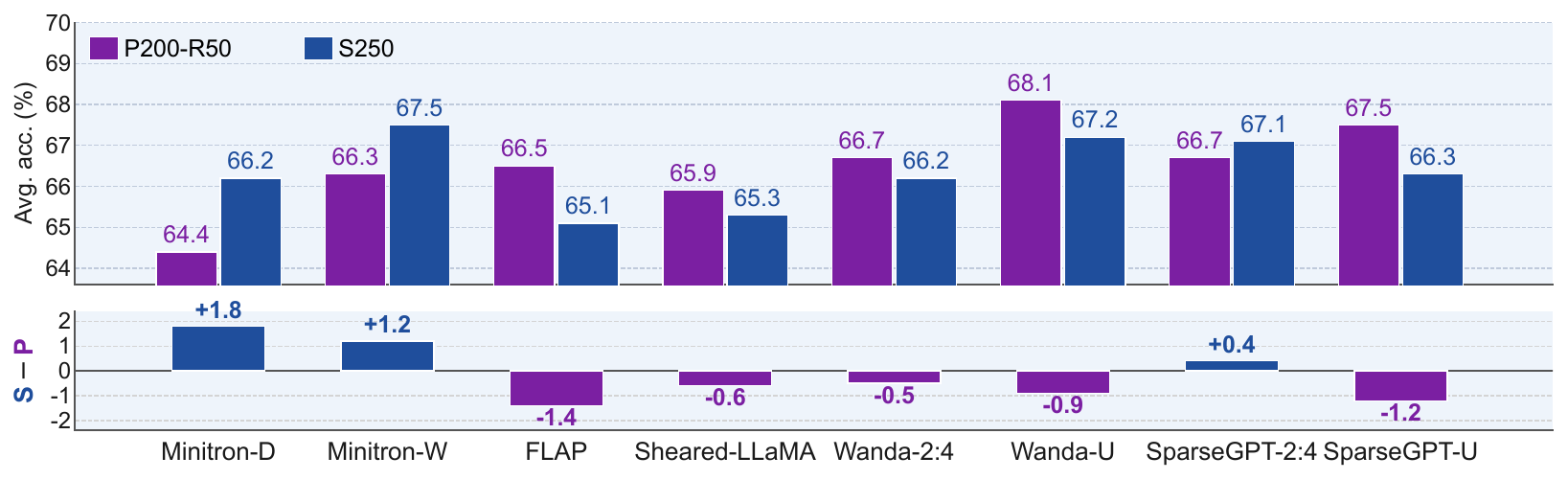}
    \caption{\textbf{P200-R50 versus S250 across pruning methods.} We plot the average accuracy (WinoGrande, ARC-C, ARC-E, HellaSwag, PIQA, SciQ, BoolQ, OBQA) from Table~\ref{table:pruning-vs-scratch-50pct}. Blue bars denote \Sx{S250} scratch training, and purple bars denote \Px{P200}-\Rx{R50} pruning followed by retraining, matching the color convention in Figure~\ref{fig:teaser}. The bottom panel shows \Px{P200}-\Rx{R50} minus \Sx{S250}; positive values indicate that pruning remains better after matching the total token budget.}
    \label{fig:p200-s250-bar}
\end{figure*}

\paragraph{Longer scratch training can match pruning at coarser granularities but not at finer ones.}
Figure~\ref{fig:p200-s250-bar} compares P200-R50 against S250 (equal total token budget). On Minitron-D, S250 surpasses P200-R50 on every benchmark (0.40 lower WikiText-2 perplexity; 0.6--4.0\% higher accuracy). The margin is smaller for the finer Minitron-W (e.g., the OBQA gain shrinks from 4.0\% to 2.2\%), showing that the advantage of coarse-granularity pruning can be recovered simply by training longer from scratch.

On sparse pruning methods, however, prolonged training from scratch only performs comparably with or worse than the pruning pipeline. For Wanda-2:4, S250 has lower perplexity than P200-R50, yet underperforms P200-R50 on 6 of 8 accuracy benchmarks, including 1.8\% and 1.2\% degradation on Winogrande and ARC Challenge respectively. On Wanda-U, S250 underperforms P200-R50 on 6 of 8 benchmarks, except BoolQ and OpenBookQA (improvements of 0.4\% and 0.8\%). This suggests that at 2:4 semi-structured sparsity training from scratch reaches parity with pruning, while at unstructured (the finest) sparsity pruning initialization remains beneficial even with matched total data. Per-token-budget numbers are tabulated in Tables~\ref{table:minitron-depth-token-scaling}--\ref{table:sheared-token-scaling} of Appendix~\ref{appendix:token-scaling}.

\subsection{Finer Granularity Unlocks Greater Benefits}
\label{subsec:granularity}

\begin{finding}
\textbf{Finding 3.} At the same pruning ratio, finer pruning granularity yields a stronger initialization and a larger advantage over training from scratch, at the cost of hardware efficiency.
\end{finding}

\paragraph{Under 50\% pruning ratio, finer pruning granularity gives the pruning pipeline more advantage.}
From Table~\ref{table:pruning-vs-scratch-50pct}, we can observe that for structured pruning methods, Minitron-W in general gives P200-R50 over S50 more advantage compared to Minitron-D; for sparse pruning methods, unstructured sparsity provides a stronger pruned model than 2:4 sparsity. For P200-R50 and S250, longer training from scratch surpasses pruning-retraining on structured methods, is roughly on par at 2:4 sparsity, and remains behind unstructured sparse pruning. Taken together, these findings suggest that under the same pruning ratio, finer granularity yields a stronger pruned model for retraining, one that outperforms the same architecture trained from scratch. These observations correspond to previous observations that finer-grained pruning methods are more capable of preserving the base model's performance, no matter whether retraining is performed or not~\citep{an2023flap, sun2024wanda, sreenivas2024minitron2}. We provide a mechanistic discussion of why sparse pruning remains uniquely hard to match by training from scratch in Appendix~\ref{appendix:sparse-advantage}.

\paragraph{The superior performance of sparse pruning is at the price of efficiency.} Table~\ref{tab:efficiency-comparison} (Appendix~\ref{appendix:efficiency-comparison}) shows a comparison between training efficiency and WikiText-103 performance of the methods we explored. All step times are profiled on Google TPU v4-256, which lacks sparse tensor cores. On this hardware, both unstructured 50\% sparse and 2:4 sparse models run at virtually the same speed as the dense model (1.0$\times$).\footnote{On NVIDIA GPUs with sparse tensor cores (e.g., A100, H100), 2:4 sparsity yields meaningful inference speedups via cuSPARSELt acceleration~\citep{mishra2021acceleratingsparse}, whereas unstructured sparsity does not. This hardware dependence does not affect the training-time comparison reported here, but practitioners targeting GPU inference should account for it.} Minitron-D achieves the highest training speedup under the same pruning ratio but at the cost of higher perplexity, while Minitron-W offers a middle ground. These observations imply that for practitioners who want compact, general-purpose models that are efficient on commodity hardware, structured pruning is the practical choice, yet it is precisely this setting where the pruning pipeline fails to outperform training from scratch under the full token budget, motivating the need for more token-efficient training paradigms.

\vspace{-1.0em}

\section{Related Work}
\label{sec:related-work}

Our work draws on three lines of research; we summarize them here and extend in Appendix~\ref{appendix:related-work}.

\paragraph{Scaling under data constraints.} As models and datasets grow to hundreds of billions of parameters and trillions of tokens~\citep{deepseekv3, qwen3, penedo2024fineweb, li2024datacomplm}, data availability rather than compute increasingly bottlenecks pretraining, motivating more token-efficient recipes~\citep{kaplan2020scalinglaw, kim2025pretraininginfinitecompute, muennighoff2025scalingdataconstrainedlm}. This is exactly the regime our study targets.

\paragraph{Continual pretraining.} Our retraining stage is effectively a form of continual pretraining (CPT)~\citep{sun2020ernie2}, where \textit{catastrophic forgetting}~\citep{mccloskey1989catastrophicinterference} and careful learning-rate re-warming are central concerns~\citep{gupta2023howtorewarmyourmodel, grattafiori2024llama3, ibrahim2024simple}. These works inform the training recipe used in our experiments.

\paragraph{Understanding network pruning.} A line of work studies whether pruned subnetworks can be matched by training from scratch, from the Lottery Ticket Hypothesis~\citep{frankle2019LTH, evci2020riggingthelottery} to pruning-scaling laws~\citep{xu2024initializingmodelswithlargerones, chen2025p2scalinglaw}. Most directly, \citet{liu2019rethinkingvaluenetworkpruning} found that for structured pruning, training the target architecture from scratch can match pruning and fine-tuning, framing pruning as a form of neural architecture search. We carry this question into the LLM regime which has far larger scale, costlier parents, and a sharper need for small deployable models. We find the answer hinges on pruning granularity. 

\paragraph{} Extended discussion of these findings and the limitations of our study are deferred to Appendix~\ref{appendix:discussion-limitations}.

\vspace{-1.0em}

\section{Conclusion}

Treating pruning as an initialization strategy rather than a compression tool, we ran a token-matched comparison against training from scratch. Under an equal training token budget, pruning initialization wins, with the margin shrinking as the pruning ratio grows. Under an equal total token budget, the outcome splits along granularity: a longer scratch run overtakes structured pruning, but sparse pruning keeps its lead.

\section*{Acknowledgments}

This work was primarily supported by the computational resources generously provided by Google's TPU Research Cloud program. We gratefully acknowledge the use of the Neuronic GPU computing cluster maintained by the Department of Computer Science at Princeton University. This work was substantially performed using Princeton Research Computing resources, a consortium led by the Princeton Institute for Computational Science and Engineering (PICSciE) and Research Computing at Princeton University.

\bibliography{main}
\bibliographystyle{plainnat}

\newpage
\appendix
\onecolumn
{\huge\bfseries Appendix}\par
\vspace{1em}
This appendix provides additional methodology details, configurations, and per-benchmark results supporting the main paper:
\begin{itemize}[noitemsep,topsep=0pt,parsep=0pt,partopsep=0pt,leftmargin=1.5em]
    \item \S~\ref{appendix:discussion-limitations} extends the conclusion with a discussion of what transfers from the larger model and enumerates the axes of variation our study leaves unexplored.
    \item \S~\ref{appendix:related-work} gives the full related-work discussion condensed in the main paper.
    \item \S~\ref{appendix:pruning-methods} catalogues the pruning methods we evaluate, with formal scoring and reconstruction objectives for each.
    \item \S~\ref{appendix:training-config} documents the training recipe, calibration sets, pruning ratios, and the architecture-search procedure used to pick a single configuration per method.
    \item \S~\ref{appendix:evaluation} specifies the perplexity corpora, downstream benchmarks, and the standard-deviation estimator we use to mark within-noise differences.
    \item \S~\ref{appendix:efficiency-comparison} reports training-speed comparisons of the pruning methods we evaluate.
    \item \S~\ref{appendix:sparse-advantage} discusses why sparse pruning remains hard to match by training from scratch, drawing on prior work on lottery tickets and weight importance.
    \item \S~\ref{appendix:detailed-results} gives the full per-benchmark breakdown, including pruning-only baselines, pruning ratio ablations, retraining-token budgets, and the equal-total-token-budget setting.
\end{itemize}
\begingroup
\hypersetup{linkcolor=black}
\etoctoccontentsline{part}{}
\etocstandardlines
\etocsettocstyle{}{}
\etocsetnexttocdepth{subsection}
\localtableofcontents
\endgroup
\vspace{1em}

\section{Extended Discussion and Limitations}
\label{appendix:discussion-limitations}

\paragraph{Discussion.} The split between sparse and structured pruning suggests that what transfers from the larger model is \emph{weight-level} information rather than architectural shape: when sparse pruning preserves the original parameter values, the surviving weights carry knowledge that extra training tokens alone do not reproduce, while a smaller dense architecture trained for longer can match a structurally pruned counterpart. This implies an asymmetric recipe: structured pruning is justified only when the training token budget is the binding constraint; otherwise a longer scratch run on a smaller dense model is competitive. Sparse pruning is the regime where the detour through a larger model genuinely pays off, but it is also the regime with the weakest hardware support---unstructured sparsity sees no speedup on the TPUs we used and only partial speedup on GPUs with sparse tensor cores. Closing this gap will likely require either better token-efficient recipes for structured small models, or broader hardware/kernel support for unstructured sparsity. A practical corollary is that the right pipeline depends on what is scarce: when target-model training data is the bottleneck, pruning beats a scratch run of equal length; when the full pipeline token budget is accounted for, only sparse pruning continues to justify the detour through the larger parent. This is a useful decision rule for practitioners, and a reminder that headline pruning numbers reported under fixed retraining budgets may overstate the practical benefit once the cost of the parent is amortized in.

\paragraph{Limitations.} Our study leaves five axes of variation unexplored. (1) \textit{Model family.} All experiments use Llama-3.1-8B as the base; we do not test whether the same pruning-vs-scratch ordering holds for other architectures (e.g.\ Qwen~\citep{qwen3}, Gemma~\citep{gemma3}, OLMo~\citep{olmo3}) or for substantially smaller and larger base models. (2) \textit{Pretraining corpus.} Both the P200 checkpoint and all retraining runs use DCLM. We do not vary the corpus, so the gap between sparse and structured pruning may shift under different data distributions such as FineWeb-Edu~\citep{penedo2024fineweb} or Dolma~\citep{soldaini2024dolma}, and the Meta-R$N$ comparisons control for architecture but not for the original Meta pretraining mixture. (3) \textit{Knowledge-distillation baselines.} We compare pruning to plain language-modeling retraining; we do not include the post-pruning knowledge-distillation pipelines used by recent structured methods (e.g.\ Minitron, ShearedLLaMA), which could narrow or close the gap for structured pruning under matched total data. (4) \textit{Data order.} Each setting is trained with a single shuffle of the retraining corpus; we do not measure run-to-run variance or the sensitivity of the pruning advantage to data ordering, which can be non-trivial in low-token regimes. (5) \textit{Random-initialization strategies.} Our scratch baselines use a single standard initialization; alternative schemes such as $\mu$P, depth-aware variance scaling, or warm-starting from a smaller pretrained checkpoint may shift the scratch curve and could change which pruning methods retain an advantage over training from scratch.

\section{Extended Related Work}
\label{appendix:related-work}

This section gives the full treatment of the three lines of research summarized in Section~\ref{sec:related-work}.

\paragraph{Model scaling and constraints.}
The traditional language model scaling paradigm assumes that simple scaling of model size and training data results in stronger performance~\citep{kaplan2020scalinglaw}, however, as recent language models scale to hundreds of billions of parameters~\citep{deepseekv3, qwen3, gemini3} and training data scales to trillions of tokens~\citep{penedo2024fineweb, li2024datacomplm}, data availability has become a practical challenge in scaling language model pretraining. Recent works on pretraining under limited data and unlimited compute have explored the effect of traditional scaling method in this scenario.~\citet{kim2025pretraininginfinitecompute} show that existing approaches such as increasing epoch and parameter count eventually overfit and improved training recipes are needed for better scaling performance.~\citet{muennighoff2025scalingdataconstrainedlm} observed similar plateau of scaling training epochs and proposed a compute optimality scaling law. Overall, these recent advances verify the practicality of our scenario where data availability is the main bottleneck in language model pretraining and more token-efficient training paradigms must be explored.

\paragraph{Continual pretraining.} As the domain-specific capabilities of LLMs become increasingly strong, \textit{continual pretraining} (CPT)~\citep{sun2020ernie2} has emerged as an approach to adapt general-purpose language models for domain specific task. The primary challenge in CPT is \textit{catastrophic forgetting}~\citep{mccloskey1989catastrophicinterference, luo2025catastrophicforgetting}, where the model loses its prior knowledge and capabilities during the continual learning stage, and the learning rate schedule must be carefully designed to mitigate this phenomenon.~\citet{gupta2023howtorewarmyourmodel} observe that starting CPT with a high learning rate causes a sharp drop in model's performance, and rewarming the LR to 10--20\% of the max learning rate significantly mitigates this problem.~\citet{grattafiori2024llama3} also suggested that learning rate schedule with re-warmup has been adopted to improve stability when continual pretraining on a new domain.~\citet{ibrahim2024simple} provide a systematic study of data replay and learning rate re-warming strategies in CPT, showing that mixing a small fraction of original pretraining data prevents catastrophic forgetting. These works provide valuable reference for the design of training recipes in our experiments, since network pruning damages the internal structure of the base model and the retraining stage is effectively a form of CPT.

\paragraph{Understanding network pruning.}
Since network pruning enables efficient acquisition of compact and capable models, its properties and impacts on the pruned models' behaviors have been a concerning topic in the network pruning community. The Lottery Ticket Hypothesis~\citep{frankle2019LTH} conjectures that every randomly initialized network contains a sparse subnetwork that can be trained to match the full network's performance; crucially, this subnetwork must be trained from its \emph{original} initialization, since random reinitialization substantially degrades performance.~\citet{evci2020riggingthelottery} further show that static sparse training from scratch can get stuck in poor local minima, whereas dynamic sparse topologies help escape them. Together, these results suggest that a sparse mask derived from a pretrained model is difficult to match via random reinitialization.~\citet{xu2024initializingmodelswithlargerones} observe that initializing small models from large pretrained models with rule-based weight selection yields better training performance.~\citet{liu2019rethinkingvaluenetworkpruning} show that small models with the same architecture trained from scratch have on-par or stronger performance than their pruned-and-retrained counterparts for structured methods, and interpret network pruning as a form of Neural Architecture Search (NAS).~\citet{chen2025p2scalinglaw} propose a pruning-scaling law connecting training data quality and post-pruning budget to final performance. These prior works have greatly inspired the experimental design of this work.

\section{Pruning Methods}
\label{appendix:pruning-methods}

\subsection{Overview}
\label{appendix:pruning-methods-overview}

Table~\ref{tab:pruning-method-taxonomy} summarizes representative LLM pruning methods organized by pruning granularity. We group methods into four categories: \textit{depth} pruning removes entire transformer layers; \textit{width} pruning reduces the hidden dimension, FFN intermediate size, or number of attention heads; \textit{depth+width} pruning jointly compresses along both axes; and \textit{sparse} pruning zero-masks individual weights within linear modules while keeping the macro-architecture intact. For each method, we list the modules it targets and the scoring metric it uses to decide which units to prune. This taxonomy motivates the six representative methods evaluated in our experiments (Minitron-depth, Minitron-width, FLAP, Sheared LLaMA, Wanda, and SparseGPT), which together span all four granularity regimes.

\begin{table}[ht!]
\centering
\arrayrulecolor{black}
\resizebox{1.0\textwidth}{!}{
\begin{tabular}{lll}
\specialrule{1.2pt}{3pt}{5pt}
\textbf{Pruning Method} & \textbf{Target Modules} & \textbf{Pruning Metric} \\
\specialrule{1.2pt}{3pt}{0pt}

\rowcolor{bestcellpurple} \multicolumn{3}{l}{\textbf{$\rightarrow$ \textit{Depth}}} \\
ShortGPT~\citep{men2024shortgpt} & layers & 
$1 - \mathbb{E}_{h,t} \dfrac{h_{i,t}^T h_{i+1,t}}{\|h_{i,t}\|_2 \|h_{i+1,t}\|_2}$ \vspace{5pt}\\

Minitron-depth~\citep{sreenivas2024minitron2} & layers &
$\mathcal{L}(\theta,z) - \mathcal{L}(\theta)$ \vspace{5pt}\\

Shortened Llama~\citep{kim2024shortenedllama} & layers &
$\left| \dfrac{\partial \mathcal{L}(D)}{\partial W_{i,j}^{k,n}} W_{i,j}^{k,n} \right|$ \\
\specialrule{1.2pt}{3pt}{0pt}

\rowcolor{bestcellpurple} \multicolumn{3}{l}{\textbf{$\rightarrow$ \textit{Width}}} \\
LLM-Pruner~\citep{ma2023llmpruner} & hidden size, FFN size, attention heads &
$\left| 
\dfrac{\partial \mathcal{L}(\mathcal{D})}{\partial W_i^k} W_i^k
-\dfrac{1}{2} \sum_{i=1}^{N} 
\left( \dfrac{\partial \mathcal{L}(\mathcal{D})}{\partial W_i^k} W_i^k \right)^2
+ \mathcal{O}\left(\|W_i^k\|^3\right)
\right|$ \vspace{5pt}\\

Minitron-width~\citep{sreenivas2024minitron2} & hidden size, FFN size &
$\|X_{:,j}\|^2,\quad 1\le j\le d$ \vspace{10pt}\\

Wanda-sp~\citep{an2023flap} & hidden size, FFN size &
$\sum_i |W_{ij}| \cdot \|X_{:,j}\|_2$ \vspace{5pt}\\

FLAP~\citep{an2023flap} & hidden size, FFN size &
$\dfrac{1}{N-1} \sum_{n=1}^{N}(X_{n,j}^l-\bar{X}_{:,j}^l)^2 \cdot \|W_{:,j}^l\|_2^2$ \\

SliceGPT~\citep{ashkboos2024slicegpt} & hidden size &
$\lambda_j\bigl(\mathbf{X}^{l\top}\mathbf{X}^{l}\bigr),\quad 1\le j\le d$ \vspace{5pt}\\
\specialrule{1.2pt}{3pt}{0pt}

\rowcolor{bestcellpurple} \multicolumn{3}{l}{\textbf{$\rightarrow$ \textit{Depth + Width}}} \\
Sheared Llama~\citep{xia2024shearedllama} & layers, hidden size, FFN size, attention heads &
$\mathcal{L}(\theta,z)
+ \sum_{j=1}^{L_S} \tilde{\mathcal{L}}_j^{\text{head}}
+ \sum_{j=1}^{L_S} \tilde{\mathcal{L}}_j^{\text{int}}
+ \tilde{\mathcal{L}}^{\text{layer}}
+ \tilde{\mathcal{L}}^{\text{hidden}}$ \\
\specialrule{1.2pt}{3pt}{0pt}

\rowcolor{bestcellpurple} \multicolumn{3}{l}{\textbf{$\rightarrow$ \textit{Sparse}}} \\
Magnitude~\citep{han2015magnitudepruning} & individual weights &
$|W_{ij}|$ \\

Sparse GPT~\citep{frantar2023sparsegpt} & individual weights &
$\dfrac{|W|^2}{\mathrm{diag}[(X^T X + \lambda I)^{-1}]}$ \vspace{5pt}\\

Wanda~\citep{sun2024wanda} & individual weights &
$|W_{ij}| \cdot \|X_j\|_2$ \\
\specialrule{1.2pt}{3pt}{5pt}
\end{tabular}
}

\caption{A taxonomy of different LLM pruning approaches categorized by pruning granularity. Depth pruning removes layers, width pruning prunes attention heads / hidden / FFN dimensions, and sparse pruning zero-masks individual weights without changing the model shape.}
\label{tab:pruning-method-taxonomy}
\end{table}

\subsection{Method details}
\label{appendix:pruning-method-details}

\paragraph{Preliminaries.} Let $\theta \in \mathbb{R}^d$ denote the parameters of a pretrained LLM and $\mathcal{D}_{\mathrm{cal}}$ a small calibration corpus. Pruning at ratio $r$ produces a smaller model by selecting a binary mask $m \in \{0,1\}^d$ that retains a fraction $1-r$ of the prunable components --- whole layers, channels, or individual weights. Existing LLM pruning methods fall into two families based on how $m$ is chosen.

\emph{(i) Loss / reconstruction minimization.} These methods select $m$ to minimize an objective on $\mathcal{D}_{\mathrm{cal}}$,
\begin{equation}
m^{\star} \;=\; \arg\min_{m \in \mathcal{M}_r}\; \mathcal{L}\bigl(\theta \odot m;\,\mathcal{D}_{\mathrm{cal}}\bigr),
\label{eq:prune-loss-family}
\end{equation}
where $\mathcal{M}_r$ is the feasible mask set at ratio $r$ and $\mathcal{L}$ is either the next-token validation loss (Minitron-depth, Sheared LLaMA) or a layerwise output-reconstruction error (SparseGPT).

\emph{(ii) Activation-based importance scoring.} These methods assign each prunable component $i$ a scalar score $s_i$ summarizing its contribution to the output activations on $\mathcal{D}_{\mathrm{cal}}$, and retain the top-$k$ highest-scoring components with $k = \lceil (1-r)\,d \rceil$:
\begin{equation}
m_i^{\star} \;=\; \mathbb{1}\!\left[\,i \in \mathrm{TopK}\bigl(\{s_j\}_{j=1}^{d},\,k\bigr)\,\right].
\label{eq:prune-score-family}
\end{equation}
The unit indexed by $i$ may be a channel (Minitron-width, FLAP) or an individual weight (Wanda).

\paragraph{Minitron-depth}~\citep{sreenivas2024minitron2} prunes a contiguous block of $k$ transformer layers, with $k$ determined by the pruning ratio. For each candidate starting index $i$, the validation loss is recomputed with layers $[i,\,i+k)$ replaced by identity mappings, and the block causing the smallest loss increase is removed:
\begin{equation}
i^{\star} \;=\; \arg\min_{i}\;\; \mathcal{L}\bigl(\theta_{[i:i+k]\to\mathrm{id}};\, \mathcal{D}_{\mathrm{cal}}\bigr) \;-\; \mathcal{L}(\theta;\, \mathcal{D}_{\mathrm{cal}}).
\end{equation}

\paragraph{Minitron-width}~\citep{sreenivas2024minitron2} scores each hidden or FFN channel $j$ by the L2 norm of its block-output activations on $\mathcal{D}_{\mathrm{cal}}$, and keeps the top-scoring channels up to the target dimension:
\begin{equation}
s_j \;=\; \bigl\| X_{:,j} \bigr\|_{2}, \qquad 1 \leq j \leq d.
\end{equation}

\paragraph{FLAP}~\citep{an2023flap} scores each hidden / FFN channel by the variance of its input feature, weighted by the squared norm of the corresponding output-weight column:
\begin{equation}
s_j \;=\; \frac{1}{N-1}\sum_{n=1}^{N}\bigl(X_{n,j} - \bar X_{:,j}\bigr)^{2} \cdot \bigl\|W_{:,j}\bigr\|_{2}^{2}.
\end{equation}
Channels with the largest $s_j$ are retained.

\paragraph{Sheared LLaMA}~\citep{xia2024shearedllama} introduces continuous mask variables $z$ over layers, attention heads, hidden channels, and FFN channels, and learns them jointly with the model parameters via a constrained optimization. The target-architecture constraints are relaxed into an augmented Lagrangian penalty:
\begin{equation}
\min_{\theta,\,z}\;\; \mathcal{L}\bigl(\theta \odot z;\,\mathcal{D}\bigr) \;+\; \sum_{c \in \mathcal{C}} \lambda_c\bigl(\hat s_c(z) - s_c^{\star}\bigr) \;+\; \phi_c\bigl(\hat s_c(z) - s_c^{\star}\bigr)^{2},
\end{equation}
where $c$ ranges over architectural axes, $\hat s_c(z)$ is the current size along axis $c$, $s_c^{\star}$ is the target, and $(\lambda_c,\phi_c)$ are dual variables updated jointly with $z$.

\paragraph{Wanda}~\citep{sun2024wanda} scores each individual weight $W_{ij}$ by its magnitude times the L2 norm of the corresponding input feature on $\mathcal{D}_{\mathrm{cal}}$, and keeps the top-scoring weights within each output row:
\begin{equation}
s_{ij} \;=\; |W_{ij}| \cdot \bigl\|X_{:,j}\bigr\|_{2}.
\end{equation}
Restricting the comparison to within each output row allows Wanda to support both unstructured and 2:4 sparsity patterns without any additional weight updates after pruning.

\paragraph{SparseGPT}~\citep{frantar2023sparsegpt} formulates pruning as a layerwise reconstruction problem. For each linear layer with weight $W$ and calibration input $X$, it jointly searches for a binary mask $m$ and updated weights $\hat W$ that minimize the output mismatch:
\begin{equation}
\min_{m,\,\hat W}\;\; \bigl\|\, X W^{\top} - X(\hat W \odot m)^{\top}\, \bigr\|_{F}^{2} \quad \text{s.t.}\quad \|m\|_{0} \leq (1-r)\,|W|.
\end{equation}
The combinatorial mask is approximated by an OBS-style score $|W_{ij}|^{2} / [(X^{\top}\!X + \lambda I)^{-1}]_{jj}$, and surviving weights are updated through the inverse Hessian to compensate for those removed.

\section{Implementation Details}
\label{appendix:training-config}

\subsection{Training recipe}
\label{appendix:training-recipe}

All pretraining and retraining experiments use the default Lingua~\citep{meta2024lingua} recipe.
Key hyperparameters are summarised in Table~\ref{tab:training-recipe}.

\begin{table}[h]
    \centering
    {\fontsize{9}{11}\selectfont
    \begin{tabular}{ll}
         \toprule
         config & value  \\
         \midrule
         sequence length & 8192 \\
         batch size & 512 \\
         weight decay & 0.01 \\
         warmup ratio & 5\% \\
         learning rate schedule & cosine decay \\
         min learning rate ratio & 0.1 \\
         optimizer & AdamW \\
         optimizer momentum & $\beta_1$ = 0.9, $\beta_2$ = 0.95 \\
         optimizer eps & 1e-8 \\
         \bottomrule
    \end{tabular}
    }
    \vspace{5pt}
    \caption{Training recipes of our pretraining and retraining experiments.}
    \label{tab:training-recipe}
\end{table}

\paragraph{Learning rate schedule.} A cosine decay schedule is used for all runs.
The warmup phase occupies 5\% of total steps, after which the learning rate decays to 10\% of its peak value.
For each experiment we sweep the peak learning rate over $\{$\texttt{1e-5, 3e-5, 1e-4, 3e-4, 1e-3}$\}$ using a short trial prefix and select the value that yields the lowest training loss.

\subsection{Pruning configurations}
\label{appendix:pruning-details}

We describe the calibration sets and configurations for each pruning method.

\paragraph{Minitron-depth.} Calibration set: 1024 samples of 8192 tokens from WikiText-103, used to compute validation loss. Pruning ratios $\{50\%, 62.5\%, 75\%, 81.3\%\}$ correspond to 16, 8, 4, and 2 hidden layers in the pruned model.

\paragraph{Minitron-width.} Same calibration set as Minitron-depth. L2-norm of block output activations is used to score each hidden channel and MLP channel. Pruning ratios $\{50\%, 62.5\%, 75\%, 81.3\%\}$. Since Minitron-width prunes hidden-size and MLP-size dimensions independently, multiple target architectures are possible for a given parameter budget. To select the final configuration at each pruning ratio, we enumerate several candidate dimension pairs (hidden size, MLP size) that yield roughly the same total parameter count, following the original Minitron paper in keeping attention heads fixed at 32 and constraining $2h \le \text{MLP} \le 4h$. We prune the pretrained Llama-3.1-8B checkpoint to each candidate and evaluate validation loss on the calibration set without retraining. The candidate with the \textbf{lowest validation loss} is used for all subsequent retraining. The selected configurations are listed in Table~\ref{tab:model-configs}.

\paragraph{FLAP.} Calibration set: 128 samples of 8192 tokens from C4. Channels are scored by input feature variance weighted by the corresponding weight column norm, and pruned at a 50\% ratio.

\paragraph{Sheared LLaMA.} We specify a target architecture matching a 50\% pruning ratio and jointly learn pruning masks over layers, attention heads, hidden dimensions, and feedforward dimensions on the training data.

\paragraph{SparseGPT and Wanda.} Calibration set: 128 samples of 8192 tokens from C4. Both 2:4 semi-structured and 50\% unstructured sparsity patterns are applied.

\subsection{Architecture search}
\label{appendix:minitron-width-nas}

For \method{Minitron-W}, \method{FLAP}, and \method{Sheared LLaMA}, the pruning method does not uniquely determine the output architecture for a given parameter budget: multiple dimension configurations can yield the same total parameter count.
For each method we enumerate a set of candidate architectures that match the target budget, prune the pretrained Llama-3.1-8B checkpoint to each candidate without any retraining, and evaluate validation loss on the calibration set.
The candidate with the \textbf{lowest validation loss} is selected and used for all subsequent retraining experiments in this paper.

\paragraph{Minitron-W.} Following the original Minitron paper, attention heads are kept unchanged (32 heads). We vary the hidden size and constrain MLP size such that $2h \le \text{MLP} \le 4h$. The full candidate set and selected configuration are listed in Table~\ref{tab:nas-minitron-w}.

\paragraph{FLAP.} Following the UL-MM configuration in the FLAP paper, hidden size is kept unchanged at 4096. We vary the number of KV heads and MLP size to reach the target of $\sim$4.5B parameters, consistent with \method{Minitron}, \method{Wanda}, and \method{SparseGPT}. Candidates and the selected configuration are listed in Table~\ref{tab:nas-flap}.

\paragraph{Sheared LLaMA.} We vary attention heads, hidden size, and MLP size to reach $\sim$4.5B parameters. Because varying all three dimensions simultaneously yields too many combinations, we fix hidden size to two representative values (3072 and 3456) and vary attention heads and MLP size within each. The full candidate set and selected configuration are listed in Table~\ref{tab:nas-sheared-llama}.

\begin{table}[!t]
    \centering
    \setlength{\tabcolsep}{4pt}
    {\fontsize{8.8}{10.8}\selectfont
    \begin{tabular}{cccccc}
        \toprule
        Ratio & Hidden & Attn heads & MLP & Params & Sel. \\
        \midrule
        \multirow{5}{*}{62.5\%}
        & 2176 & 32 & 7168 & 3.1B & \\
        & 2304 & 32 & 6656 & 3.1B & \\
        & \cellcolor{bestcellpurple}2432 & \cellcolor{bestcellpurple}32 & \cellcolor{bestcellpurple}6144 & \cellcolor{bestcellpurple}3.1B & \cellcolor{bestcellpurple}$\checkmark$ \\
        & 2560 & 32 & 6400 & 3.1B & \\
        & 2688 & 32 & 5888 & 3.1B & \\
        \midrule
        \multirow{4}{*}{75\%}
        & 1664 & 32 & 6656 & 2.0B & \\
        & \cellcolor{bestcellpurple}1792 & \cellcolor{bestcellpurple}32 & \cellcolor{bestcellpurple}6016 & \cellcolor{bestcellpurple}2.0B & \cellcolor{bestcellpurple}$\checkmark$ \\
        & 1920 & 32 & 5248 & 2.0B & \\
        & 2048 & 32 & 4480 & 2.0B & \\
        \midrule
        \multirow{3}{*}{81.3\%}
        & 1408 & 32 & 5632 & 1.5B & \\
        & \cellcolor{bestcellpurple}1536 & \cellcolor{bestcellpurple}32 & \cellcolor{bestcellpurple}4736 & \cellcolor{bestcellpurple}1.5B & \cellcolor{bestcellpurple}$\checkmark$ \\
        & 1664 & 32 & 3840 & 1.5B & \\
        \bottomrule
    \end{tabular}
    }
    \caption{Candidate architectures for \method{Minitron-W}. Attention heads are fixed at 32; we vary hidden size with $2h \le \text{MLP} \le 4h$. The selected configuration at each pruning ratio is marked with $\checkmark$.}
    \label{tab:nas-minitron-w}
\end{table}

\begin{table}[!t]
    \centering
    \setlength{\tabcolsep}{4pt}
    {\fontsize{8.8}{10.8}\selectfont
    \begin{tabular}{cccccc}
        \toprule
        Ratio & Hidden & KV heads & MLP & Params & Sel. \\
        \midrule
        \multirow{5}{*}{50\%}
        & 4096 & 16 & 7168 & 4.5B & \\
        & \cellcolor{bestcellpurple}4096 & \cellcolor{bestcellpurple}20 & \cellcolor{bestcellpurple}6656 & \cellcolor{bestcellpurple}4.5B & \cellcolor{bestcellpurple}$\checkmark$ \\
        & 4096 & 24 & 6272 & 4.5B & \\
        & 4096 & 28 & 5888 & 4.5B & \\
        & 4096 & 32 & 5376 & 4.5B & \\
        \bottomrule
    \end{tabular}
    }
    \caption{Candidate architectures for \method{FLAP}. Following the UL-MM config in the FLAP paper, hidden size is fixed at 4096; KV heads and MLP size are varied. The selected configuration is marked with $\checkmark$.}
    \label{tab:nas-flap}
\end{table}

\begin{table}[!t]
    \centering
    \setlength{\tabcolsep}{4pt}
    {\fontsize{8.8}{10.8}\selectfont
    \begin{tabular}{cccccc}
        \toprule
        Ratio & Hidden & Attn heads & MLP & Params & Sel. \\
        \midrule
        \multirow{10}{*}{50\%}
        & 3072 & 16 & 11008 & 4.5B & \\
        & 3072 & 20 & 10496 & 4.5B & \\
        & 3072 & 24 & 10112 & 4.5B & \\
        & 3072 & 28 & 9728 & 4.5B & \\
        & 3072 & 32 & 9216 & 4.5B & \\
        & 3456 & 16 & 9216 & 4.5B & \\
        & 3456 & 20 & 8832 & 4.5B & \\
        & 3456 & 24 & 8448 & 4.5B & \\
        & 3456 & 28 & 7936 & 4.5B & \\
        & \cellcolor{bestcellpurple}3456 & \cellcolor{bestcellpurple}32 & \cellcolor{bestcellpurple}7552 & \cellcolor{bestcellpurple}4.5B & \cellcolor{bestcellpurple}$\checkmark$ \\
        \noalign{\vspace{-3pt}}\bottomrule
    \end{tabular}
    }
    \caption{Candidate architectures for \method{Sheared LLaMA}. Hidden size is fixed to 3072 or 3456; attention heads and MLP size are varied within each. The selected configuration is marked with $\checkmark$.}
    \label{tab:nas-sheared-llama}
\end{table}

\section{Evaluation protocol}
\label{appendix:evaluation}

\paragraph{Linguistic perplexity.} We evaluate on the general-domain corpora C4~\citep{raffel2020c4}, WikiText-103~\citep{merity2017wikitext}, and WikiText-2~\citep{merity2017wikitext}, along with the news-and-summaries corpus CNN Dailymail~\citep{chen2016cnndailymail}. For each corpus, we collect 256 sequences of length 8192 (the max position embedding of Llama-3.1) as the evaluation set.

\paragraph{Downstream accuracy.} We evaluate on eight zero-shot benchmarks spanning three categories: commonsense QA (WinoGrande~\citep{sakaguchi2020winogrande}, HellaSwag~\citep{zellers2019hellaswag}, PIQA~\citep{bisk2019piqa}), scientific QA (ARC-Challenge~\citep{clark2018arc}, ARC-Easy~\citep{clark2018arc}, SciQ~\citep{welbl2017sciq}, OpenBookQA~\citep{mihaylov2018obqa}), and logical reasoning (BoolQ~\citep{clark2019boolq}). Zero-shot evaluation measures foundational knowledge and capability, independent of in-context learning biases. The random-chance baselines are 50\% for WinoGrande (binary), 25\% for ARC-C and ARC-E (four-way), and 50\% for PIQA and BoolQ (binary); results near these baselines indicate that the model has essentially lost the measured capability, and we interpret them accordingly.

\paragraph{Evaluation framework and standard deviations.} All downstream evaluations are run with lm-evaluation-harness~\citep{eleuther2021lmeval}. Grey entries in Table~\ref{table:pruning-vs-scratch-50pct} denote differences that fall within the evaluation standard deviation, estimated as $\sqrt{acc(1-acc)/N}$ where $N$ is the number of evaluation examples. The per-benchmark standard deviations are shown in Table~\ref{table:eval-std}.

\begin{table}[!t]
\centering
\begin{tabular}{lc}
\toprule
\textbf{Benchmark} & \textbf{Std.\ Dev.} \\
\midrule
WinoG & $\pm$1.3\% \\
ARC-C & $\pm$1.4\% \\
ARC-E & $\pm$1.0\% \\
HSwag & $\pm$0.4\% \\
PIQA  & $\pm$1.0\% \\
SciQ  & $\pm$0.8\% \\
BoolQ & $\pm$0.8\% \\
OBQA  & $\pm$2.2\% \\
\bottomrule
\end{tabular}
\caption{Per-benchmark evaluation standard deviations.}
\label{table:eval-std}
\end{table}

\section{Efficiency comparison}
\label{appendix:efficiency-comparison}

\begin{table}[hbt!]
    \centering
    {\fontsize{9}{11}\selectfont
    \begin{tabular}{lccc}
        \toprule
         method & sparsity & FLOPs (TF) & Speedup \\
         \midrule\noalign{\vspace{-3pt}}
         base & 0.0\% & $32.9$ & 1.0$\times$ \\
         \noalign{\vspace{-3pt}}\midrule
         \multirow{4}{*}{Minitron-D}
         & 50.0\% & $17.5$ & \textcolor{Good}{1.9$\times$} \\
         & 62.5\% & $9.8$ & \textcolor{Good}{4.2$\times$} \\
         & 75.0\% & $6.0$ & \textcolor{Good}{7.2$\times$} \\
         & 81.3\% & $4.1$ & \textcolor{Good}{11.2$\times$} \\
         \midrule
         \multirow{4}{*}{Minitron-W}
         & 50.0\% & $19.1$ & \textcolor{Good}{1.4$\times$} \\
         & 62.5\% & $12.6$ & \textcolor{Good}{2.0$\times$} \\
         & 75.0\% & $9.8$ & \textcolor{Good}{2.4$\times$} \\
         & 81.3\% & $7.9$ & \textcolor{Good}{2.9$\times$} \\
         \midrule
         FLAP & 50.0\% & $17.7$ & \textcolor{Good}{1.5$\times$} \\
         \midrule
         Sheared LLaMA & 50.0\% & $18.9$ & \textcolor{Good}{1.4$\times$} \\
         \midrule
         \multirow{2}{*}{Wanda}
         & 50.0\% & $16.5$ & \textcolor{Bad}{1.0$\times$} \\
         & 2:4 & $16.5$ & \textcolor{Bad}{1.0$\times$} \\
         \midrule
         \multirow{2}{*}{SparseGPT}
         & 50\% & $16.5$ & \textcolor{Bad}{1.0$\times$} \\
         & 2:4 & $16.5$ & \textcolor{Bad}{1.0$\times$} \\
         \bottomrule
    \end{tabular}
    }
    \caption{\textbf{Efficiency comparison between pruning methods.} Models are obtained by pretraining Llama-3.1-8B for 200B tokens, pruning at the listed ratio, and retraining for 50B tokens. For Wanda and SparseGPT, 50\% denotes unstructured sparsity. FLOPs are computed for a single forward pass with sequence length 2048; for sparse models, theoretical FLOPs assume 50\% of weight multiplications are skipped. \textbf{All step times are profiled on Google TPU v4-256 using MaxText~\citep{aihypercomputer2025maxtext}, which has no sparse tensor cores.} On this hardware, sparse methods yield no training speedup despite halved theoretical FLOPs. On NVIDIA GPUs with sparse tensor cores (A100/H100), 2:4 sparse models obtain meaningful speedups via cuSPARSELt; unstructured sparse models do not.}
    \label{tab:efficiency-comparison}
\end{table}

Table~\ref{tab:efficiency-comparison} compares training speed and WikiText-103~\citep{merity2017wikitext} performance among the pruning methods we experiment with. Minitron-D~\citep{sreenivas2024minitron2} with the coarsest granularity provides the most speedup while preserving the least of the base model's performance on WikiText-103, while sparse pruning methods like SparseGPT~\citep{frantar2023sparsegpt} and Wanda~\citep{sun2024wanda} score the best but provide no training speedup at all. Minitron-W~\citep{sreenivas2024minitron2} provides a middle ground with moderate speedup and decent preservation of the base model's performance.

An overview of the methods is shown in Figure~\ref{fig:prune-methods}.

Among these methods, we cover four different granularities: depth pruning, width pruning, 2:4 sparse pruning, and unstructured sparse pruning. We also use four pruning ratios 50\%, 62.5\%, 75.0\% and 81.3\%.

\section{Why Does Sparse Pruning Maintain Its Advantage?}
\label{appendix:sparse-advantage}

The persistent advantage of sparse pruning over training from scratch, even when scratch training uses the full 250B-token budget (Section~\ref{subsec:granularity}), calls for a mechanistic explanation. We offer two complementary perspectives grounded in prior work.

\paragraph{Optimizing a sparse model from scratch is hard.}
The Lottery Ticket Hypothesis~\citep{frankle2019LTH} posits that a pruned subnetwork trained from its \emph{original} initialization converges faster and to higher accuracy than the same sparse structure trained from a \emph{random} reinitialization.~\citet{evci2020riggingthelottery} further demonstrate that static sparse training can get stuck in isolated local minima, and that allowing the sparse topology to evolve during training helps escape them. In our S250-sparse setting, the sparse mask is fixed and the weights are randomly reinitialized, placing it in exactly the hard regime identified by these works.

\paragraph{Sparse pruning identifies and preserves non-redundant weights.}
\citet{liu2019rethinkingvaluenetworkpruning} observe that both fine-tuned and scratch-trained pruned models have far fewer near-zero weights than the unpruned base model, suggesting that pruning actively identifies structurally important weights. For unstructured sparse pruning in particular, the fine-tuned model diverges more strongly from the scratch-trained counterpart than structured pruning does, indicating that the specific weight \emph{values} inherited from the large model, not just the sparsity pattern, are a key source of advantage.

\section{Detailed Per-Benchmark Results}
\label{appendix:detailed-results}

\subsection{Pruning-only baselines}
\label{appendix:pruning-only}

To contextualize the benefit of retraining, we report zero-shot performance immediately after pruning, before any retraining is applied.
Table~\ref{tab:pruning-only} covers both structured pruning (Minitron-D, Minitron-W, FLAP, Sheared LLaMA) and sparse pruning (Wanda and SparseGPT at 2:4 and unstructured 50\% sparsity).
We note that some structured-pruning configurations exhibit extremely high post-pruning perplexity --- for example, P200 + Minitron-W reaches 24686.5 on C4 and 36761.5 on WikiText-2 --- because activation-norm channel selection on the P200 checkpoint can leave the residual stream poorly conditioned before any weights are re-tuned. Despite this near-broken starting point, the same configuration fully recovers after a modest retraining budget: at the 50B-token retraining setting reported in the main text (Table~\ref{table:pruning-vs-scratch-50pct}), P200-R50 with Minitron-W reaches accuracy and perplexity comparable to the other 50\%-pruned methods, indicating that initial post-pruning perplexity alone is a poor predictor of the retrained model's final quality.

\begin{table*}[h]
    \centering
    \setlength{\tabcolsep}{3.5pt}
    {\footnotesize
    \begin{tabular}{llrrccccccccc}
        \specialrule{1.2pt}{3pt}{5pt}
        \multirow{2}{*}{Base model} & \multirow{2}{*}{Method}
        & \multicolumn{2}{c}{perplexity $\downarrow$} & \multicolumn{9}{c}{accuracy(\%) $\uparrow$} \\
        \cmidrule(lr){3-4} \cmidrule(lr){5-13}
        & & C4 & WT-2 & WinoG & ARC-C & ARC-E & HSwag & PIQA & SciQ & BoolQ & OBQA & \textbf{Avg} \\
        \specialrule{1.2pt}{3pt}{0pt}
        \rowcolor{bestcellpurple} Meta & Base & 9.9 & 5.8 & 74.3 & 53.4 & 81.1 & 78.9 & 81.3 & 96.3 & 82.0 & 45.0 & 74.0 \\
        Meta & \makecell[l]{Minitron-D} & 532.4 & 599.9 & 48.5 & 24.0 & 30.6 & 29.1 & 54.7 & 31.1 & 37.9 & 27.2 & 35.4 \\
        Meta & \makecell[l]{Minitron-W} & 411.0 & 182.0 & 52.4 & 22.6 & 30.3 & 28.6 & 53.0 & 51.2 & 37.8 & 27.4 & 37.9 \\
        Meta & FLAP & 85.4 & 34.2 & 53.4 & 25.5 & 36.8 & 34.6 & 57.7 & 61.9 & 38.8 & 30.0 & 42.3 \\
        Meta & \makecell[l]{Sheared-LLaMA} & 19.1 & 13.3 & 61.6 & 35.0 & 59.3 & 57.0 & 69.9 & 88.9 & 65.7 & 36.2 & 59.2 \\
        Meta & Wanda 2:4 & 38.0 & 21.6 & 59.4 & 30.5 & 52.0 & 49.2 & 69.0 & 83.8 & 67.9 & 30.8 & 55.3 \\
        Meta & \makecell[l]{Wanda-U} & 15.5 & 9.3 & 70.4 & 42.2 & \best{68.3} & 68.5 & 76.4 & \best{90.8} & 79.1 & 40.2 & 67.0 \\
        Meta & \makecell[l]{SparseGPT-2:4} & 24.2 & 15.0 & 63.5 & 34.8 & 59.3 & 56.5 & 70.5 & 87.3 & 68.2 & 35.0 & 59.4 \\
        Meta & \makecell[l]{SparseGPT-U} & \best{14.5} & \best{8.8} & \best{72.5} & \best{43.0} & 67.4 & \best{71.4} & \best{76.5} & 87.9 & \best{79.2} & \best{42.4} & \best{67.5} \\
        \specialrule{1.2pt}{3pt}{0pt}
        \rowcolor{bestcellpurple} P200 & Base & 15.8 & 9.0 & 68.4 & 46.3 & 75.1 & 74.5 & 78.8 & 95.3 & 69.3 & 41.0 & 68.6 \\
        P200 & \makecell[l]{Minitron-D} & 504.0 & 364.2 & 53.6 & 29.9 & 34.7 & 36.9 & 58.8 & 35.4 & 53.8 & 28.6 & 41.5 \\
        P200 & \makecell[l]{Minitron-W} & 24686.5 & 36761.5 & 49.3 & 26.3 & 25.8 & 26.2 & 51.2 & 22.8 & 38.9 & 27.6 & 33.5 \\
        P200 & FLAP & 128.0 & 57.0 & 51.5 & 24.8 & 47.8 & 38.1 & 63.9 & 69.4 & 62.1 & 31.2 & 48.6 \\
        P200 & \makecell[l]{Sheared-LLaMA} & \best{20.4} & 15.2 & 61.1 & 41.8 & 69.4 & 64.1 & 74.9 & \best{91.5} & 65.8 & 37.6 & 63.3 \\
        P200 & Wanda-2:4 & 142.8 & 67.1 & 57.5 & 30.5 & 53.9 & 52.7 & 67.6 & 89.5 & 63.9 & 32.0 & 56.0 \\
        P200 & \makecell[l]{Wanda-U} & 30.9 & 15.0 & 63.4 & 42.4 & 67.8 & \best{69.1} & 76.4 & 90.3 & \best{66.9} & 38.2 & 64.3 \\
        P200 & \makecell[l]{SparseGPT-2:4} & 42.7 & 20.3 & 62.6 & 35.8 & 63.5 & 59.4 & 73.5 & 89.1 & 65.6 & 37.4 & 60.9 \\
        P200 & \makecell[l]{SparseGPT-U} & 22.4 & \best{11.9} & \best{66.5} & \best{43.9} & \best{70.2} & 68.9 & \best{77.9} & 90.4 & 65.4 & \best{40.2} & \best{65.4} \\
        \bottomrule
    \end{tabular}
    }
    \caption{Zero-shot performance immediately after pruning, without any retraining. All pruning methods use 50\% sparsity. \textit{Minitron-D} and \textit{Minitron-W} denote the depth and width variants of Minitron pruning, respectively. \textit{Wanda-U} and \textit{SparseGPT-U} denote the unstructured sparsity variants of Wanda and SparseGPT (as opposed to the 2:4 semi-structured variants). C4 and WT-2: perplexity (lower is better); others: 0-shot accuracy (\%). Avg is the mean over all 8 zero-shot benchmarks (WinoG, ARC-C, ARC-E, HSwag, PIQA, SciQ, BoolQ, OBQA). Purple-shaded rows mark the unpruned base model in each block (Meta's released Llama-3.1-8B and our 200B-token pretrained checkpoint); \textcolor{brandpurple}{\textbf{purple bold values}} mark the best result among the pruned methods within each block.}
    \label{tab:pruning-only}
\end{table*}

\subsection{Pruning ratio ablations}
\label{appendix:pruning-ratio}

Table~\ref{table:pruning-ratio-ablations} reports per-benchmark results for Minitron-D and Minitron-W at pruning ratios $\{50.0\%,\allowbreak 62.5\%,\allowbreak 75.0\%,\allowbreak 81.3\%\}$ (equal training token budget). The $\Delta$P200-R50 rows summarize the advantage of pruning initialization over scratch training, which diminishes as the pruning ratio grows.

\begin{table*}[!t]
\centering
\renewcommand{\up}[1]{\cellcolor{diffbetter}+#1}%
\renewcommand{\down}[1]{\cellcolor{diffworse}-#1}%
\renewcommand{\same}[1]{\cellcolor{diffneutral}#1}%
\begin{adjustbox}{max width=\textwidth}
{\large
\setlength{\tabcolsep}{2.2pt}
\newcommand{\tblm}{10pt}
\begin{tabular}{@{\hspace{\tblm}}llrrrrrrrrrrrrrr@{\hspace{\tblm}}}
\toprule
\multirow{2}{*}{ratio} & \multirow{2}{*}{tokens}
& \multicolumn{1}{c}{loss $\downarrow$} & \multicolumn{4}{c}{perplexity $\downarrow$} & \multicolumn{9}{c}{accuracy(\%) $\uparrow$}\\
\cmidrule{3-3} \cmidrule(lr){4-7} \cmidrule(lr){8-16}
& & DCLM & C4 & WT & WT-2 & CNN &  WinoG & ARC-C & ARC-E & HSwag & PIQA & SciQ & BoolQ & OBQA & \textbf{Avg} \\
\specialrule{1.2pt}{3pt}{0pt}
\rowcolor{bestcellpurple}
\multicolumn{16}{l}{\textbf{$\rightarrow$ Minitron-\textit{depth}}} \\
\multirow{3}{*}{\makecell[l]{50.0\%}}
& S50 & 2.64 & 16.90 & 10.61 & 10.77 & 10.35 & 56.8 & 36.2 & 70.1 & 61.9 & 74.4 & 90.4 & 57.8 & 37.8 & 60.7 \\[3pt]
& P200-R50 & 2.54 & 15.34 & 9.30 & 9.41 & 9.46 & 64.4 & 40.7 & 71.5 & 68.0 & 76.8 & 91.7 & 64.0 & 38.0 & 64.4
\\[-1pt]
& {\footnotesize $\Delta$P200-R50} & \up{0.10} & \up{1.56} & \up{1.31} & \up{1.36} & \up{0.89} & \up{7.6} & \up{4.5} & \up{1.4} & \up{6.1} & \up{2.4} & \up{1.3} & \up{6.2} & \up{0.2} & \up{3.7} \\
\noalign{\vspace{-3pt}}\midrule
\multirow{3}{*}{\makecell[l]{62.5\%}}
& S50 & 2.82 & 20.23 & 13.43 & 13.70 & 12.57 & 54.0 & \textbf{32.3} & 62.3 & 51.1 & 71.4 & 85.2 & 51.6 & 33.8 & 55.2 \\[3pt]
& P200-R50 & 2.77 & 19.35 & 12.67 & 12.89 & 12.03 & 55.3 & 30.7 & 62.5 & 53.6 & 72.3 & \textbf{87.3} & 55.7 & 34.2 & 56.4
\\[-1pt]
& {\footnotesize $\Delta$P200-R50} & \up{0.05} & \up{0.88} & \up{0.76} & \up{0.81} & \up{0.53} & \up{1.3} & \down{1.6} & \up{0.2} & \up{2.5} & \up{0.9} & \up{2.1} & \up{4.1} & \up{0.4} & \up{1.2} \\
\noalign{\vspace{-3pt}}\midrule
\multirow{3}{*}{\makecell[l]{75.0\%}}
& S50 & 3.02 & 24.57 & 17.29 & 17.69 & 15.55 & 51.4 & 27.6 & 56.2 & 40.9 & 68.1 & 79.5 & 55.0 & 29.0 & 51.0 \\[3pt]
& P200-R50 & 3.01 & 24.41 & 17.11 & 17.48 & 15.49 & 51.1 & 29.3 & 55.8 & 41.4 & 68.6 & 81.9 & 54.9 & 30.4 & 51.7
\\[-1pt]
& {\footnotesize $\Delta$P200-R50}  & \same{0.01} & \up{0.16} & \up{0.18} & \up{0.21} & \up{0.06} & \down{0.3} & \up{1.7}& \down{0.4} & \up{0.6} & \up{0.5} & \up{2.4} & \down{0.2} & \up{1.4} & \up{0.7} \\
\noalign{\vspace{-3pt}}\midrule
\multirow{3}{*}{\makecell[l]{81.3\%}}
& S50 & 3.25 & 30.79 & 24.57 & 25.16 & 19.97 & 49.4 & 25.9 & 52.2 & 35.7 & 65.1 & 74.4 & 55.1 & 28.8 & 48.3 \\[3pt]
& P200-R50 & 3.25 & 30.79 & 24.92 & 25.48 & 20.33 & 51.2 & 24.6 & 53.5 & 35.4 & 66.4 & 76.0 & 50.4 & 27.4 & 48.1
\\[-1pt]
& {\footnotesize $\Delta$P200-R50} & \same{0.00} & \same{0.00} & \down{0.35} & \down{0.32} & \down{0.36} & \up{1.8} & \down{1.4} & \up{1.2} & \down{0.3} & \up{1.3} & \up{1.6} & \down{4.7} & \down{1.4} & \down{0.2} \\
\noalign{\vspace{-3pt}}\specialrule{1.2pt}{3pt}{0pt}
\rowcolor{bestcellpurple}
\multicolumn{16}{l}{\textbf{$\rightarrow$ Minitron-\textit{width}}} \\
\multirow{3}{*}{\makecell[l]{50.0\%}}
& S50 & 2.70 & 17.47 & 10.99 & 11.17 & 10.65 & 61.3 & 33.6 & 65.1 & 60.3 & 74.7 & 89.6 & 60.8 & 33.8 & 59.9 \\[3pt]
& P200-R50 & 2.47 & 14.6 & 8.93 & 9.01 & 8.92 & 65.4 & 41.8 & 73.4 & 70.5 & 77.4 & 93.8 & 67.4 & 40.4 & 66.3
\\[-1pt]
& {\footnotesize $\Delta$P200-R50} & \up{0.23} & \up{2.87} & \up{2.05} & \up{2.16} & \up{1.73} & \up{4.1} & \up{8.2} & \up{8.3} & \up{10.2} & \up{2.7} & \up{4.2} & \up{6.6} & \up{6.6} & \up{6.4} \\
\noalign{\vspace{-3pt}}\midrule
\multirow{3}{*}{\makecell[l]{62.5\%}}
& S50 & 2.75 & 18.67 & 12.00 & 12.23 & 11.45 & 56.1 & 30.1 & 65.2 & 56.3 & 72.9 & 86.2 & 55.7 & 32.6 & 56.9 \\[3pt]
& P200-R50 & 2.50 & 14.51 & 8.49 & 8.61 & 9.01 & 61.5 & 41.7 & 71.0 & 69.5 & 75.4 & 92.6 & 72.4 & 35.4 & 64.9
\\[-1pt]
& {\footnotesize $\Delta$P200-R50} & \up{0.24} & \up{4.16} & \up{3.51} & \up{3.61} & \up{2.44} & \up{5.4} & \up{11.6} & \up{5.8} & \up{13.2} & \up{2.4} & \up{6.4} & \up{16.8} & \up{2.8} & \up{8.0} \\
\noalign{\vspace{-3pt}}\midrule
\multirow{3}{*}{\makecell[l]{75.0\%}}
& S50 & 2.82 & 20.00 & 12.99 & 13.26 & 12.27 & 56.3 & 29.8 & 61.7 & 52.6 & 71.9 & 87.4 & 55.7 & 32.2 & 56.0 \\[3pt]
& P200-R50 & 2.60 & 15.98 & 9.61 & 9.76 & 9.91 & 57.2 & 37.2 & 68.0 & 63.5 & 73.4 & 89.8 & 66.1 & 32.4 & 61.0
\\[-1pt]
& {\footnotesize $\Delta$P200-R50} & \up{0.21} & \up{4.03} & \up{3.38} & \up{3.49} & \up{2.36} & \up{0.9} & \up{7.4} & \up{6.2} & \up{11.0} & \up{1.5} & \up{2.4} & \up{10.4} & \up{0.2} & \up{5.0} \\
\noalign{\vspace{-3pt}}\midrule
\multirow{3}{*}{\makecell[l]{81.3\%}}
& S50 & 2.86 & 20.93 & 13.77 & 14.08 & 12.88 & 54.1 & 27.9 & 60.7 & 49.5 & 70.7 & 86.0 & 53.2 & 31.5 & 54.2 \\[3pt]
& P200-R50 & 2.63 & 16.16 & 9.53 & 9.71 & 10.10 & 54.0 & 35.8 & 65.8 & 61.2 & 71.7 & 88.3 & 67.7 & 29.8 & 59.3
\\[-1pt]
& {\footnotesize $\Delta$P200-R50} & \up{0.23} & \up{4.77} & \up{4.24} & \up{4.37} & \up{2.78} & \down{0.1} & \up{7.9} & \up{5.1} & \up{11.7} & \up{1.0} & \up{2.3} & \up{14.5} & \down{1.7} & \up{5.1} \\
\noalign{\vspace{-3pt}}\bottomrule
\end{tabular}
}
\end{adjustbox}
\caption{\textbf{Effect of pruning ratio on the initialization advantage (equal training token budget).} \textit{P200-R50:} pretrain 200B tokens, prune, retrain 50B tokens. \textit{S50:} train from scratch for 50B tokens. $\Delta$P200-R50: difference relative to P200-R50. Blue: P200-R50 better; red: P200-R50 worse; grey: within evaluation standard deviation. The advantage of pruning initialization diminishes as the pruning ratio increases, vanishing near 81.3\% for depth pruning.}
\label{table:pruning-ratio-ablations}
\end{table*}

\paragraph{Per-benchmark observations.} BoolQ is the most volatile benchmark across pruning ratios under both granularities: under depth pruning it decays monotonically from $+6.2$ at 50\% to $-4.7$ at 81.3\%, while under width pruning it swings non-monotonically between $+6.6$ and $+16.8$. ARC-Challenge under Minitron-D 62.5\% is a localized anomaly --- pruning costs $1.6$ points there even though neighboring ratios (50\%, 75\%) both favor pruning. Minitron-W retains a positive average advantage at every ratio we tested (including 81.3\%, $\Delta\text{Avg}=+5.1$), in sharp contrast to depth pruning, where the average advantage vanishes at 81.3\% and only WinoG and ARC-E remain positive.

\subsection{Token scaling across pruning methods}
\label{appendix:token-scaling}

Tables~\ref{table:minitron-depth-token-scaling},~\ref{table:minitron-width-token-scaling},~\ref{table:flap-token-scaling}, and~\ref{table:sheared-token-scaling} report per-benchmark results across retraining token budgets $\{10\text{B},\allowbreak 30\text{B},\allowbreak 50\text{B},\allowbreak 250\text{B},\allowbreak 500\text{B}\}$ for Minitron-D, Minitron-W, FLAP, and Sheared LLaMA, respectively, under the three initialization strategies (train from scratch, prune from our 200B-pretrained checkpoint, and prune from Meta's released Llama-3.1-8B). All four tables are the full-data sources behind Figure~\ref{fig:minitron-token-scaling} in the main text.

\begin{table*}[!t]
\renewcommand{\up}[1]{\cellcolor{diffbetter}+#1}%
\renewcommand{\down}[1]{\cellcolor{diffworse}-#1}%
\renewcommand{\same}[1]{\cellcolor{diffneutral}#1}%
\centering
\begin{adjustbox}{max width=\textwidth}
{\large
\setlength{\tabcolsep}{2.2pt}
\newcommand{\tblm}{10pt}
\begin{tabular}{@{\hspace{\tblm}}llrrrrrrrrrrrrrr@{\hspace{\tblm}}}
\toprule
\multirow{2}{*}{method} & \multirow{2}{*}{tokens}
& \multicolumn{1}{c}{loss $\downarrow$} & \multicolumn{4}{c}{perplexity $\downarrow$} & \multicolumn{9}{c}{accuracy(\%) $\uparrow$}\\
\cmidrule{3-3} \cmidrule(lr){4-7} \cmidrule(lr){8-16}
& & DCLM & C4 & WT & WT-2 & CNN & WinoG & ARC-C & ARC-E & HSwag & PIQA & SciQ & BoolQ & OBQA & \textbf{Avg} \\
\midrule
\multirow{25}{*}{\cellcolor{white}\makecell[l]{Minitron\\depth}}
& S10 & 3.09 & 25.92 & 18.18 & 18.71 & 16.07 & 49.0 & 26.7 & 49.3 & 38.6 & 67.2 & 76.7 & 60.0 & 31.2 & 49.8 \\
& P200-R10 & 2.70 & 16.83 & 10.29 & 10.44 & 10.79 & 57.7 & 34.6 & 68.5 & 60.3 & 75.8 & 89.8 & 60.5 & 38.2 & 60.7
\\[-1pt]
& {\footnotesize $\Delta$S10} & \up{0.39} & \up{9.09} & \up{7.89} & \up{8.27} & \up{5.28} & \up{8.7} & \up{7.9} & \up{19.2} & \up{21.7} & \up{8.6} & \up{13.1} & \up{0.5} & \up{7.0} & \up{10.9} \\
& Meta-R10 & 2.69 & 16.29 & 9.86 & 9.98 & 10.58 & 58.5 & 36.7 & 70.1 & 61.9 & 75.6 & 92.0 & 61.2 & 39.4 & 61.9
\\[-1pt]
& {\footnotesize $\Delta$S10} & \up{0.41} & \up{9.63} & \up{8.32} & \up{8.73} & \up{5.49} & \up{9.5} & \up{10.0} & \up{20.8} & \up{23.3} & \up{8.4} & \up{15.3} & \up{1.2} & \up{8.2} & \up{12.1}
\\[3pt]
& S30 & 2.81 & 19.40 & 12.50 & 12.75 & 11.91 & 54.3 & 30.4 & 63.7 & 54.2 & 72.3 & 86.1 & 48.8 & 33.8 & 55.5 \\
& P200-R30 & 2.59 & 15.40 & 9.19 & 9.27 & 9.68 & 59.0 & 38.3 & 71.2 & 65.6 & 76.5 & 91.2 & 61.5 & 37.8 & 62.6
\\[-1pt]
& {\footnotesize $\Delta$S30} & \up{0.21} & \up{4.00} & \up{3.31} & \up{3.48} & \up{2.23} & \up{4.7} & \up{7.9} & \up{7.5} & \up{11.4} & \up{4.2} & \up{5.1} & \up{12.7} & \up{4.0} & \up{7.1} \\
& Meta-R30 & 2.59 & 15.07 & 8.96 & 9.04 & 9.64 & 61.7 & 39.8 & 72.8 & 66.8 & 76.7 & 93.0 & 61.2 & 40.6 & 64.1
\\[-1pt]
& {\footnotesize $\Delta$S30} & \up{0.22} & \up{4.33} & \up{3.54} & \up{3.71} & \up{2.27} & \up{7.4} & \up{9.4} & \up{9.1} & \up{12.6} & \up{4.4} & \up{6.9} & \up{12.4} & \up{6.8} & \up{8.6}
\\[3pt]
& S50 & 2.64 & 16.90 & 10.61 & 10.77 & 10.35 & 56.8 & 36.2 & 70.1 & 61.9 & 74.4 & 90.4 & 57.8 & 37.8 & 60.7 \\
& P200-R50 & 2.54 & 15.34 & 9.30 & 9.41 & 9.46 & 64.4 & 40.7 & 71.5 & 68.0 & 76.8 & 91.7 & 64.0 & 38.0 & 64.4
\\[-1pt]
& {\footnotesize $\Delta$S50} & \up{0.10} & \up{1.56} & \up{1.31} & \up{1.36} & \up{0.89} & \up{7.6} & \up{4.5} & \up{1.4} & \up{6.1} & \up{2.4} & \up{1.3} & \up{6.2} & \up{0.2} & \up{3.7} \\
& Meta-R50 & 2.50 & 14.42 & 8.55 & 8.61 & 9.21 & 63.3 & 40.0 & 72.7 & 69.0 & 77.6 & 93.2 & 62.2 & 40.6 & 63.4
\\[-1pt]
& {\footnotesize $\Delta$S50} & \up{0.14} & \up{2.48} & \up{2.06} & \up{2.16} & \up{1.14} & \up{6.5} & \up{3.8} & \up{2.6} & \up{7.1} & \up{3.2} & \up{2.8} & \up{4.4} & \up{2.8} & \up{2.7}
\\[3pt]
& S250 & 2.49 & 14.78 & 8.93 & 9.01 & 9.03 & 63.5 & 41.7 & 73.7 & 70.8 & 78.0 & 93.2 & 65.1 & 42.0 & 66.0 \\
& P200-R250 & 2.46 & 13.87 & 8.12 & 8.16 & 8.57 & 67.3 & 43.3 & 74.2 & 72.5 & 78.6 & 94.4 & 62.7 & 42.8 & 67.0
\\[-1pt]
& {\footnotesize $\Delta$S250} & \up{0.03} & \up{0.91} & \up{0.81} & \up{0.85} & \up{0.46} & \up{3.8} & \up{1.6} & \up{0.5} & \up{1.7} & \up{0.6} & \up{1.2} & \down{2.4} & \up{0.8} & \up{1.0} \\

& Meta-R250 & 2.46 & 13.72 & 8.00 & 8.03 & 8.56 & 66.6 & 44.5 & 75.0 & 72.9 & 78.8 & 94.4 & 68.6 & 43.0 & 68.0
\\[-1pt]
& {\footnotesize $\Delta$S250} & \up{0.03} & \up{1.06} & \up{0.93} & \up{0.98} & \up{0.47} & \up{3.1} & \up{2.8} & \up{1.3} & \up{2.1} & \up{0.8} & \up{1.2} & \up{3.5} & \up{1.0} & \up{2.0}
\\[3pt]
& S500 & 2.47 & 14.15 & 8.42 & 8.46 & 8.65 & 68.4 & 42.7 & 74.6 & 72.5 & 77.9 & 93.4 & 68.1 & 40.8 & 67.3 \\
& P200-R500 & 2.43 & 13.59 & 7.93 & 7.96 & 8.36 & 66.4 & 45.0 & 74.6 & 74.1 & 79.0 & 94.6 & 68.5 & 42.4 & 68.1
\\[-1pt]
& {\footnotesize $\Delta$S500} & \up{0.04} & \up{0.56} & \up{0.49} & \up{0.50} & \up{0.29} & \down{2.0} & \up{2.3} & \same{0.0} & \up{1.6} & \up{1.1} & \up{1.2} & \up{0.4} & \up{1.6} & \up{0.8} \\
& Meta-R500 & 2.43 & 13.46 & 7.81 & 7.83 & 8.34 & 68.7 & 45.8 & 75.7 & 74.5 & 80.0 & 95.0 & 69.4 & 44.4 & 69.2
\\[-1pt]
& {\footnotesize $\Delta$S500} & \up{0.04} & \up{0.69} & \up{0.61} & \up{0.63} & \up{0.31} & \up{0.3} & \up{3.1} & \up{1.1} & \up{2.0} & \up{2.1} & \up{1.6} & \up{1.3} & \up{3.6} & \up{1.9} \\
\noalign{\vspace{-3pt}}\bottomrule
\end{tabular}
}
\end{adjustbox}
\caption{\textbf{Minitron depth pruning across retraining token budgets (Llama-3.1-8B $\to$ 4B, 50\% pruning).} \textit{S$N$:} train from scratch for $N$B tokens. \textit{P200-R$N$:} pretrain 200B tokens on DCLM, prune, retrain $N$B tokens. \textit{Meta-R$N$:} prune from Meta's released Llama-3.1-8B, retrain $N$B tokens. DCLM loss is validation cross-entropy on DCLM (lower is better). Avg is the mean of WinoG, ARC-C, ARC-E, HSwag, PIQA, SciQ, BoolQ, and OBQA. Each $\Delta$S$N$ row reports the difference between the pruned-model row directly above (P200-R$N$ or Meta-R$N$) and the S$N$ baseline.}
\label{table:minitron-depth-token-scaling}
\end{table*}

\begin{table*}[!t]
\renewcommand{\up}[1]{\cellcolor{diffbetter}+#1}%
\renewcommand{\down}[1]{\cellcolor{diffworse}-#1}%
\renewcommand{\same}[1]{\cellcolor{diffneutral}#1}%
\centering
\begin{adjustbox}{max width=\textwidth}
{\large
\setlength{\tabcolsep}{2.2pt}
\newcommand{\tblm}{10pt}
\begin{tabular}{@{\hspace{\tblm}}llrrrrrrrrrrrrrr@{\hspace{\tblm}}}
\toprule
\multirow{2}{*}{method} & \multirow{2}{*}{tokens}
& \multicolumn{1}{c}{loss $\downarrow$} & \multicolumn{4}{c}{perplexity $\downarrow$} & \multicolumn{9}{c}{accuracy(\%) $\uparrow$}\\
\cmidrule{3-3} \cmidrule(lr){4-7} \cmidrule(lr){8-16}
& & DCLM & C4 & WT & WT-2 & CNN & WinoG & ARC-C & ARC-E & HSwag & PIQA & SciQ & BoolQ & OBQA & \textbf{Avg} \\
\midrule
\multirow{25}{*}{\cellcolor{white}\makecell[l]{Minitron\\width}}
& S10 & 3.11 & 26.68 & 18.74 & 19.26 & 16.55 & 51.9 & 25.3 & 51.5 & 38.2 & 67.1 & 78.6 & 59.7 & 19.8 & 46.8 \\
& P200-R10 & 2.62 & 15.57 & 9.44 & 9.58 & 9.83 & 65.0 & 37.2 & 69.6 & 68.3 & 76.2 & 93.9 & 66.0 & 29.8 & 63.3
\\[-1pt]
& {\footnotesize $\Delta$S10} & \up{0.49} & \up{11.11} & \up{9.30} & \up{9.68} & \up{6.72} & \up{13.1} & \up{11.9} & \up{18.1} & \up{30.1} & \up{9.1} & \up{15.3} & \up{6.3} & \up{10.0} & \up{16.5} \\
& Meta-R10 & 2.55 & 14.40 & 8.35 & 8.46 & 9.07 & 66.8 & 42.2 & 73.3 & 70.9 & 78.3 & 95.1 & 73.6 & 29.6 & 66.3
\\[-1pt]
& {\footnotesize $\Delta$S10} & \up{0.56} & \up{12.28} & \up{10.39} & \up{10.80} & \up{7.48} & \up{14.9} & \up{16.9} & \up{21.8} & \up{32.7} & \up{11.2} & \up{16.5} & \up{13.9} & \up{9.8} & \up{19.5}
\\[3pt]
& S30 & 2.80 & 19.19 & 12.32 & 12.55 & 11.79 & 56.4 & 30.3 & 63.2 & 55.1 & 72.5 & 86.4 & 57.3 & 23.4 & 55.5 \\
& P200-R30 & 2.47 & 13.66 & 7.94 & 8.03 & 8.55 & 68.9 & 48.0 & 76.3 & 73.8 & 79.9 & 95.6 & 76.0 & 31.0 & 69.4
\\[-1pt]
& {\footnotesize $\Delta$S30} & \up{0.32} & \up{5.53} & \up{4.38} & \up{4.52} & \up{3.24} & \up{12.5} & \up{17.7} & \up{13.1} & \up{18.7} & \up{7.4} & \up{9.2} & \up{18.7} & \up{7.6} & \up{13.9} \\
& Meta-R30 & 2.47 & 13.46 & 7.73 & 7.80 & 8.45 & 68.7 & 45.2 & 76.3 & 74.2 & 78.7 & 95.3 & 75.8 & 32.8 & 68.6
\\[-1pt]
& {\footnotesize $\Delta$S30} & \up{0.33} & \up{5.73} & \up{4.59} & \up{4.75} & \up{3.34} & \up{12.3} & \up{14.9} & \up{13.1} & \up{19.1} & \up{6.2} & \up{8.9} & \up{18.5} & \up{9.4} & \up{13.1}
\\[3pt]
& S50 & 2.70 & 17.47 & 10.99 & 11.17 & 10.62 & 61.3 & 33.6 & 65.1 & 60.3 & 74.7 & 89.6 & 60.8 & 24.2 & 59.0 \\
& P200-R50 & 2.47 & 14.59 & 8.93 & 9.01 & 8.92 & 65.7 & 41.7 & 73.4 & 70.5 & 77.4 & 93.8 & 67.4 & 29.8 & 65.7
\\[-1pt]
& {\footnotesize $\Delta$S50} & \up{0.23} & \up{2.88} & \up{2.06} & \up{2.16} & \up{1.70} & \up{4.4} & \up{8.1} & \up{8.3} & \up{10.2} & \up{2.7} & \up{4.2} & \up{6.6} & \up{5.6} & \up{6.7} \\
& Meta-R50 & 2.39 & 12.95 & 7.41 & 7.47 & 8.13 & 69.7 & 48.5 & 77.6 & 75.7 & 79.0 & 95.8 & 75.3 & 32.3 & 70.1
\\[-1pt]
& {\footnotesize $\Delta$S50} & \up{0.31} & \up{4.52} & \up{3.58} & \up{3.70} & \up{2.49} & \up{8.4} & \up{14.9} & \up{12.5} & \up{15.4} & \up{4.3} & \up{6.2} & \up{14.5} & \up{8.1} & \up{11.1}
\\[3pt]
& S250 & 2.46 & 14.37 & 8.62 & 8.69 & 8.75 & 65.6 & 45.9 & 75.4 & 72.2 & 77.8 & 91.1 & 68.5 & 31.0 & 67.4 \\
& P200-R250 & 2.39 & 13.02 & 7.56 & 7.60 & 8.02 & 66.3 & 45.2 & 76.0 & 75.2 & 78.8 & 95.5 & 75.6 & 28.0 & 68.3
\\[-1pt]
& {\footnotesize $\Delta$S250} & \up{0.07} & \up{1.35} & \up{1.06} & \up{1.09} & \up{0.73} & \up{0.7} & \down{0.7} & \up{0.6} & \up{3.0} & \up{1.0} & \up{4.4} & \up{7.1} & \down{3.0} & \up{0.9} \\
& Meta-R250 & 2.37 & 12.51 & 7.13 & 7.16 & 7.78 & 70.6 & 48.8 & 78.7 & 77.3 & 79.8 & 96.0 & 78.3 & 32.6 & 71.1
\\[-1pt]
& {\footnotesize $\Delta$S250} & \up{0.10} & \up{1.86} & \up{1.49} & \up{1.53} & \up{0.97} & \up{5.0} & \up{2.9} & \up{3.3} & \up{5.1} & \up{2.0} & \up{4.9} & \up{9.8} & \up{1.6} & \up{3.7}
\\[3pt]
& S500 & 2.44 & 13.68 & 8.11 & 8.15 & 8.36 & 66.9 & 45.8 & 75.9 & 73.9 & 79.9 & 94.8 & 71.6 & 32.2 & 68.5 \\
& P200-R500 & 2.35 & 12.52 & 7.12 & 7.14 & 7.73 & 66.4 & 48.2 & 77.2 & 77.2 & 79.2 & 95.9 & 78.6 & 26.8 & 69.6
\\[-1pt]
& {\footnotesize $\Delta$S500} & \up{0.09} & \up{1.16} & \up{0.99} & \up{1.01} & \up{0.63} & \down{0.5} & \up{2.4} & \up{1.3} & \up{3.3} & \down{0.7} & \up{1.1} & \up{7.0} & \down{5.4} & \up{1.1} \\
& Meta-R500 & 2.34 & 12.35 & 6.99 & 7.01 & 7.65 & 72.1 & 50.6 & 79.8 & 78.1 & 80.1 & 95.7 & 79.5 & 33.0 & 72.1
\\[-1pt]
& {\footnotesize $\Delta$S500} & \up{0.09} & \up{1.33} & \up{1.12} & \up{1.14} & \up{0.71} & \up{5.2} & \up{4.8} & \up{3.9} & \up{4.2} & \up{0.2} & \up{0.9} & \up{7.9} & \up{0.8} & \up{3.6} \\
\noalign{\vspace{-3pt}}\bottomrule
\end{tabular}
}
\end{adjustbox}
\caption{\textbf{Minitron width pruning across retraining token budgets (Llama-3.1-8B $\to$ 4B, 50\% pruning).} \textit{S$N$:} train from scratch for $N$B tokens. \textit{P200-R$N$:} pretrain 200B tokens on DCLM, prune, retrain $N$B tokens. \textit{Meta-R$N$:} prune from Meta's released Llama-3.1-8B, retrain $N$B tokens. DCLM loss is validation cross-entropy on DCLM (lower is better). Avg is the mean of WinoG, ARC-C, ARC-E, HSwag, PIQA, SciQ, BoolQ, and OBQA. Each $\Delta$S$N$ row reports the difference between the pruned-model row directly above (P200-R$N$ or Meta-R$N$) and the S$N$ baseline.}
\label{table:minitron-width-token-scaling}
\end{table*}

\begin{table*}[!t]
\renewcommand{\up}[1]{\cellcolor{diffbetter}+#1}%
\renewcommand{\down}[1]{\cellcolor{diffworse}-#1}%
\renewcommand{\same}[1]{\cellcolor{diffneutral}#1}%
\centering
\begin{adjustbox}{max width=\textwidth}
{\large
\setlength{\tabcolsep}{2.2pt}
\newcommand{\tblm}{10pt}
\begin{tabular}{@{\hspace{\tblm}}llrrrrrrrrrrrrrr@{\hspace{\tblm}}}
\toprule
\multirow{2}{*}{method} & \multirow{2}{*}{tokens}
& \multicolumn{1}{c}{loss $\downarrow$} & \multicolumn{4}{c}{perplexity $\downarrow$} & \multicolumn{9}{c}{accuracy(\%) $\uparrow$}\\
\cmidrule{3-3} \cmidrule(lr){4-7} \cmidrule(lr){8-16}
& & DCLM & C4 & WT & WT-2 & CNN & WinoG & ARC-C & ARC-E & HSwag & PIQA & SciQ & BoolQ & OBQA & \textbf{Avg} \\
\midrule
\multirow{25}{*}{\cellcolor{white}\makecell[l]{FLAP}}
& S10 & 3.15 & 26.38 & 18.55 & 19.08 & 16.43 & 49.0 & 26.5 & 51.3 & 38.5 & 66.8 & 72.0 & 59.5 & 20.5 & 48.0 \\
& P200-R10 & 2.62 & 14.75 & 8.40 & 8.53 & 9.51 & 63.7 & 40.8 & 72.8 & 68.0 & 76.7 & 93.0 & 73.0 & 27.2 & 64.4
\\[-1pt]
& {\footnotesize $\Delta$S10} & \up{0.52} & \up{11.63} & \up{10.15} & \up{10.56} & \up{6.92} & \up{14.7} & \up{14.3} & \up{21.5} & \up{29.5} & \up{9.8} & \up{21.0} & \up{13.5} & \up{6.7} & \up{16.4} \\
& Meta-R10 & 2.58 & 14.40 & 8.15 & 8.28 & 9.25 & 66.0 & 43.0 & 74.5 & 70.5 & 77.8 & 93.8 & 74.5 & 30.0 & 66.3
\\[-1pt]
& {\footnotesize $\Delta$S10} & \up{0.57} & \up{11.98} & \up{10.40} & \up{10.80} & \up{7.18} & \up{17.0} & \up{16.5} & \up{23.2} & \up{32.0} & \up{11.0} & \up{21.8} & \up{15.0} & \up{9.5} & \up{18.3}
\\[3pt]
& S30 & 2.85 & 19.36 & 12.47 & 12.72 & 11.94 & 53.9 & 30.9 & 64.3 & 54.7 & 72.1 & 80.7 & 52.7 & 24.9 & 54.3 \\
& P200-R30 & 2.54 & 14.10 & 8.10 & 8.22 & 8.95 & 65.5 & 43.5 & 75.5 & 71.0 & 77.5 & 93.5 & 74.5 & 27.4 & 66.1
\\[-1pt]
& {\footnotesize $\Delta$S30} & \up{0.31} & \up{5.26} & \up{4.37} & \up{4.50} & \up{2.99} & \up{11.6} & \up{12.6} & \up{11.2} & \up{16.3} & \up{5.4} & \up{12.8} & \up{21.8} & \up{2.5} & \up{11.8} \\
& Meta-R30 & 2.51 & 13.85 & 7.95 & 8.02 & 8.85 & 67.8 & 45.8 & 75.8 & 72.5 & 78.3 & 93.8 & 74.9 & 31.0 & 67.5
\\[-1pt]
& {\footnotesize $\Delta$S30} & \up{0.34} & \up{5.51} & \up{4.52} & \up{4.70} & \up{3.09} & \up{13.9} & \up{14.9} & \up{11.5} & \up{17.8} & \up{6.2} & \up{13.1} & \up{22.2} & \up{6.1} & \up{13.2}
\\[3pt]
& S50 & 2.71 & 17.57 & 11.16 & 11.34 & 10.77 & 58.3 & 34.9 & 67.2 & 59.8 & 74.3 & 83.3 & 58.9 & 25.4 & 57.8 \\
& P200-R50 & 2.47 & 13.68 & 7.89 & 7.97 & 8.52 & 66.9 & 44.8 & 76.2 & 72.9 & 77.8 & 93.9 & 75.5 & 27.6 & 67.0
\\[-1pt]
& {\footnotesize $\Delta$S50} & \up{0.24} & \up{3.89} & \up{3.27} & \up{3.37} & \up{2.25} & \up{8.6} & \up{9.9} & \up{9.0} & \up{13.1} & \up{3.5} & \up{10.6} & \up{16.6} & \up{2.2} & \up{9.2} \\
& Meta-R50 & 2.47 & 13.43 & 7.69 & 7.75 & 8.39 & 69.2 & 46.3 & 76.5 & 74.6 & 78.8 & 93.8 & 75.3 & 31.6 & 68.3
\\[-1pt]
& {\footnotesize $\Delta$S50} & \up{0.24} & \up{4.14} & \up{3.47} & \up{3.59} & \up{2.38} & \up{10.9} & \up{11.4} & \up{9.3} & \up{14.8} & \up{4.5} & \up{10.5} & \up{16.4} & \up{6.2} & \up{10.5}
\\[3pt]
& S250 & 2.51 & 14.57 & 8.76 & 8.82 & 8.92 & 66.1 & 43.2 & 74.8 & 71.7 & 77.9 & 91.1 & 66.8 & 29.8 & 65.2 \\
& P200-R250 & 2.40 & 12.85 & 7.35 & 7.38 & 8.00 & 68.8 & 48.3 & 76.9 & 76.1 & 79.3 & 94.8 & 79.8 & 32.8 & 69.6
\\[-1pt]
& {\footnotesize $\Delta$S250} & \up{0.11} & \up{1.72} & \up{1.41} & \up{1.44} & \up{0.92} & \up{2.7} & \up{5.1} & \up{2.1} & \up{4.4} & \up{1.4} & \up{3.7} & \up{13.0} & \up{3.0} & \up{4.4} \\
& Meta-R250 & 2.39 & 12.80 & 7.30 & 7.33 & 7.93 & 69.9 & 48.5 & 77.2 & 76.4 & 79.8 & 93.9 & 75.4 & 34.2 & 69.4
\\[-1pt]
& {\footnotesize $\Delta$S250} & \up{0.12} & \up{1.77} & \up{1.46} & \up{1.49} & \up{0.99} & \up{3.8} & \up{5.3} & \up{2.4} & \up{4.7} & \up{1.9} & \up{2.8} & \up{8.6} & \up{4.4} & \up{4.2}
\\[3pt]
& S500 & 2.49 & 13.91 & 8.25 & 8.28 & 8.53 & 69.2 & 43.7 & 75.5 & 73.4 & 78.9 & 93.0 & 69.8 & 31.4 & 66.9 \\
& P200-R500 & 2.36 & 11.89 & 6.40 & 6.43 & 7.42 & 68.8 & 51.5 & 79.7 & 79.3 & 79.8 & 96.4 & 83.3 & 26.8 & 70.7
\\[-1pt]
& {\footnotesize $\Delta$S500} & \up{0.13} & \up{2.02} & \up{1.85} & \up{1.85} & \up{1.11} & \down{0.5} & \up{7.9} & \up{4.2} & \up{5.9} & \up{0.9} & \up{3.4} & \up{13.5} & \down{4.6} & \up{3.8} \\
& Meta-R500 & 2.35 & 11.75 & 6.28 & 6.31 & 7.35 & 72.3 & 52.0 & 80.3 & 80.0 & 80.4 & 95.8 & 79.5 & 36.5 & 72.1
\\[-1pt]
& {\footnotesize $\Delta$S500} & \up{0.14} & \up{2.16} & \up{1.97} & \up{1.97} & \up{1.18} & \up{3.1} & \up{8.3} & \up{4.8} & \up{6.6} & \up{1.5} & \up{2.8} & \up{9.7} & \up{5.1} & \up{5.2} \\
\noalign{\vspace{-3pt}}\bottomrule
\end{tabular}
}
\end{adjustbox}
\caption{\textbf{FLAP across retraining token budgets (Llama-3.1-8B $\to$ 4B, 50\% pruning).} \textit{S$N$:} train from scratch for $N$B tokens. \textit{P200-R$N$:} pretrain 200B tokens on DCLM, prune, retrain $N$B tokens. \textit{Meta-R$N$:} prune from Meta's released Llama-3.1-8B, retrain $N$B tokens. DCLM loss is validation cross-entropy on DCLM (lower is better). Avg is the mean of WinoG, ARC-C, ARC-E, HSwag, PIQA, SciQ, BoolQ, and OBQA.}
\label{table:flap-token-scaling}
\end{table*}

\begin{table*}[!t]
\renewcommand{\up}[1]{\cellcolor{diffbetter}+#1}%
\renewcommand{\down}[1]{\cellcolor{diffworse}-#1}%
\renewcommand{\same}[1]{\cellcolor{diffneutral}#1}%
\centering
\begin{adjustbox}{max width=\textwidth}
{\large
\setlength{\tabcolsep}{2.2pt}
\newcommand{\tblm}{10pt}
\begin{tabular}{@{\hspace{\tblm}}llrrrrrrrrrrrrrr@{\hspace{\tblm}}}
\toprule
\multirow{2}{*}{method} & \multirow{2}{*}{tokens}
& \multicolumn{1}{c}{loss $\downarrow$} & \multicolumn{4}{c}{perplexity $\downarrow$} & \multicolumn{9}{c}{accuracy(\%) $\uparrow$}\\
\cmidrule{3-3} \cmidrule(lr){4-7} \cmidrule(lr){8-16}
& & DCLM & C4 & WT & WT-2 & CNN & WinoG & ARC-C & ARC-E & HSwag & PIQA & SciQ & BoolQ & OBQA & \textbf{Avg} \\
\midrule
\multirow{25}{*}{\cellcolor{white}\makecell[l]{Sheared\\LLaMA}}
& S10 & 3.10 & 26.30 & 18.46 & 18.98 & 16.31 & 50.7 & 25.9 & 50.6 & 38.4 & 67.1 & 77.8 & 59.8 & 19.0 & 48.7 \\
& P200-R10 & 2.62 & 15.45 & 9.42 & 9.55 & 9.62 & 63.4 & 39.8 & 71.5 & 67.8 & 75.8 & 91.8 & 64.0 & 29.0 & 62.9
\\[-1pt]
& {\footnotesize $\Delta$S10} & \up{0.48} & \up{10.85} & \up{9.04} & \up{9.43} & \up{6.69} & \up{12.7} & \up{13.9} & \up{20.9} & \up{29.4} & \up{8.7} & \up{14.0} & \up{4.2} & \up{10.0} & \up{14.2} \\
& Meta-R10 & 2.58 & 14.95 & 8.95 & 9.05 & 9.30 & 65.5 & 42.5 & 73.6 & 70.2 & 77.0 & 93.2 & 67.0 & 31.0 & 65.0
\\[-1pt]
& {\footnotesize $\Delta$S10} & \up{0.52} & \up{11.35} & \up{9.51} & \up{9.93} & \up{7.01} & \up{14.8} & \up{16.6} & \up{23.0} & \up{31.8} & \up{9.9} & \up{15.4} & \up{7.2} & \up{12.0} & \up{16.3}
\\[3pt]
& S30 & 2.80 & 19.29 & 12.41 & 12.65 & 11.85 & 55.6 & 30.3 & 63.4 & 54.7 & 72.4 & 86.3 & 53.9 & 23.2 & 55.0 \\
& P200-R30 & 2.55 & 14.70 & 8.85 & 8.96 & 9.10 & 64.8 & 42.0 & 73.6 & 70.0 & 76.4 & 92.8 & 66.2 & 32.0 & 64.7
\\[-1pt]
& {\footnotesize $\Delta$S30} & \up{0.25} & \up{4.59} & \up{3.56} & \up{3.69} & \up{2.75} & \up{9.2} & \up{11.7} & \up{10.2} & \up{15.3} & \up{4.0} & \up{6.5} & \up{12.3} & \up{8.8} & \up{9.7} \\
& Meta-R30 & 2.52 & 14.25 & 8.55 & 8.65 & 8.90 & 66.0 & 43.5 & 74.6 & 71.3 & 77.2 & 94.0 & 68.0 & 33.5 & 66.0
\\[-1pt]
& {\footnotesize $\Delta$S30} & \up{0.28} & \up{5.04} & \up{3.86} & \up{4.00} & \up{2.95} & \up{10.4} & \up{13.2} & \up{11.2} & \up{16.6} & \up{4.8} & \up{7.7} & \up{14.1} & \up{10.3} & \up{11.0}
\\[3pt]
& S50 & 2.67 & 17.51 & 11.10 & 11.28 & 10.69 & 60.0 & 34.2 & 66.1 & 59.9 & 74.6 & 89.0 & 59.6 & 23.8 & 58.4 \\
& P200-R50 & 2.50 & 14.35 & 8.55 & 8.63 & 8.90 & 65.8 & 42.8 & 74.7 & 71.2 & 76.7 & 93.6 & 68.0 & 34.2 & 65.9
\\[-1pt]
& {\footnotesize $\Delta$S50} & \up{0.17} & \up{3.16} & \up{2.55} & \up{2.65} & \up{1.79} & \up{5.8} & \up{8.6} & \up{8.6} & \up{11.3} & \up{2.1} & \up{4.6} & \up{8.4} & \up{10.4} & \up{7.5} \\
& Meta-R50 & 2.47 & 13.95 & 8.25 & 8.32 & 8.70 & 67.0 & 44.6 & 75.5 & 72.5 & 77.8 & 94.3 & 69.0 & 35.0 & 67.0
\\[-1pt]
& {\footnotesize $\Delta$S50} & \up{0.20} & \up{3.56} & \up{2.85} & \up{2.96} & \up{1.99} & \up{7.0} & \up{10.4} & \up{9.4} & \up{12.6} & \up{3.2} & \up{5.3} & \up{9.4} & \up{11.2} & \up{8.6}
\\[3pt]
& S250 & 2.47 & 14.57 & 8.77 & 8.85 & 8.89 & 64.8 & 44.2 & 74.7 & 71.6 & 77.9 & 91.9 & 67.1 & 30.4 & 65.3 \\
& P200-R250 & 2.42 & 13.35 & 7.78 & 7.85 & 8.35 & 67.5 & 45.8 & 76.4 & 74.2 & 79.4 & 94.6 & 70.0 & 35.5 & 67.9
\\[-1pt]
& {\footnotesize $\Delta$S250} & \up{0.05} & \up{1.22} & \up{0.99} & \up{1.00} & \up{0.54} & \up{2.7} & \up{1.6} & \up{1.7} & \up{2.6} & \up{1.5} & \up{2.7} & \up{2.9} & \up{5.1} & \up{2.6} \\
& Meta-R250 & 2.39 & 13.00 & 7.45 & 7.52 & 8.15 & 70.8 & 49.2 & 79.0 & 77.2 & 80.5 & 95.8 & 74.0 & 37.0 & 70.4
\\[-1pt]
& {\footnotesize $\Delta$S250} & \up{0.08} & \up{1.57} & \up{1.32} & \up{1.33} & \up{0.74} & \up{6.0} & \up{5.0} & \up{4.3} & \up{5.6} & \up{2.6} & \up{3.9} & \up{6.9} & \up{6.6} & \up{5.1}
\\[3pt]
& S500 & 2.45 & 13.91 & 8.26 & 8.30 & 8.50 & 67.5 & 44.6 & 75.4 & 73.3 & 79.1 & 94.2 & 70.2 & 31.9 & 67.0 \\
& P200-R500 & 2.37 & 12.75 & 7.25 & 7.32 & 8.00 & 69.5 & 47.8 & 78.0 & 75.8 & 79.6 & 95.8 & 73.0 & 37.0 & 69.6
\\[-1pt]
& {\footnotesize $\Delta$S500} & \up{0.08} & \up{1.16} & \up{1.01} & \up{0.98} & \up{0.50} & \up{2.0} & \up{3.2} & \up{2.6} & \up{2.5} & \up{0.5} & \up{1.6} & \up{2.8} & \up{5.1} & \up{2.6} \\
& Meta-R500 & 2.34 & 12.35 & 6.95 & 7.02 & 7.75 & 72.0 & 51.0 & 80.0 & 78.2 & 79.8 & 96.2 & 76.0 & 38.5 & 71.5
\\[-1pt]
& {\footnotesize $\Delta$S500} & \up{0.11} & \up{1.56} & \up{1.31} & \up{1.28} & \up{0.75} & \up{4.5} & \up{6.4} & \up{4.6} & \up{4.9} & \up{0.7} & \up{2.0} & \up{5.8} & \up{6.6} & \up{4.5} \\
\noalign{\vspace{-3pt}}\bottomrule
\end{tabular}
}
\end{adjustbox}
\caption{\textbf{Sheared LLaMA across retraining token budgets (Llama-3.1-8B $\to$ 4B, 50\% pruning).} \textit{S$N$:} train from scratch for $N$B tokens. \textit{P200-R$N$:} pretrain 200B tokens on DCLM, prune, retrain $N$B tokens. \textit{Meta-R$N$:} prune from Meta's released Llama-3.1-8B, retrain $N$B tokens. DCLM loss is validation cross-entropy on DCLM (lower is better). Avg is the mean of WinoG, ARC-C, ARC-E, HSwag, PIQA, SciQ, BoolQ, and OBQA.}
\label{table:sheared-token-scaling}
\end{table*}

\paragraph{Per-benchmark observations.} Across all four structured methods, BoolQ is the first benchmark to flip negative as the retraining budget grows: at 500B tokens, BoolQ goes to $-2.0$ (Minitron-D), $-5.4$ (Minitron-W), and $-4.6$ (FLAP), even when other benchmarks (e.g.\ ARC-Challenge, HellaSwag) still favor pruning. This suggests BoolQ rewards raw data scale more than initialization quality. Sheared LLaMA is the only structured method whose average advantage never crosses zero across the full token range ($\Delta\text{Avg}\!\ge\!+2.6$ even at 500B), consistent with its joint-mask training-aware pruning objective. Meta-R$N$ tends to outperform P200-R$N$ at low budgets (e.g.\ Minitron-D $\Delta_{S10}\!=\!+12.1$ vs.\ $+10.9$), with the gap closing as $N$ grows --- a residual signal of the original Meta pretraining mixture beyond what DCLM-only P200 retraining recovers.

\subsection{Pruning vs.\ scratch under equal total token budget}
\label{appendix:equal-total-data}

Table~\ref{table:pruning-vs-scratch-50pct} reports the complete per-benchmark breakdown underlying Figure~\ref{fig:p200-s250-bar}, comparing P200-R50 against S50 (equal training token budget) and S250 (equal total token budget) across all six pruning methods (with both 2:4 and unstructured variants for Wanda and SparseGPT) at the 50\% pruning ratio. Under an equal total token budget (P200-R50 vs.\ S250), sparse methods retain a consistent advantage from the pruned initialization, whereas structured methods are largely matched or surpassed by extended scratch training.

\begin{table*}[!t]
\renewcommand{\up}[1]{\cellcolor{diffbetter}+#1}%
\renewcommand{\down}[1]{\cellcolor{diffworse}-#1}%
\renewcommand{\same}[1]{\cellcolor{diffneutral}#1}%
\centering
\begin{adjustbox}{max width=\textwidth}
{\large
\setlength{\tabcolsep}{2.2pt}
\newcommand{\tblm}{10pt}
\begin{tabular}{@{\hspace{\tblm}}llrrrrrrrrrrrrrr@{\hspace{\tblm}}}
\toprule
\multirow{2}{*}{method} & \multirow{2}{*}{tokens}
& \multicolumn{1}{c}{loss $\downarrow$} & \multicolumn{4}{c}{perplexity $\downarrow$} & \multicolumn{9}{c}{accuracy(\%) $\uparrow$}\\
\cmidrule{3-3} \cmidrule(lr){4-7} \cmidrule(lr){8-16}
& & DCLM & C4 & WT & WT-2 & CNN &  WinoG & ARC-C & ARC-E & HSwag & PIQA & SciQ & BoolQ & OBQA & \textbf{Avg} \\
\midrule
\multirow{5}{*}{\cellcolor{white}\makecell[l]{Minitron\\depth}}
& P200-R50 & 2.54 & 15.34 & 9.30 & 9.41 & 9.46 & 64.4 & 40.7 & 71.5 & 68.0 & 76.8 & 91.7 & 64.0 & 38.0 & 64.4 \\[3pt]
& S50 & 2.64 & 16.90 & 10.61 & 10.77 & 10.35 & 56.8 & 36.2 & 70.1 & 61.9 & 74.4 & 90.4 & 57.8 & 37.8 & 60.7
\\[-1pt]
& {\footnotesize $\Delta$P200-R50} & \down{0.10} & \down{1.56} & \down{1.31} & \down{1.36} & \down{0.89} & \down{7.6} & \down{4.5} & \down{1.4} & \down{6.1} & \down{2.4} & \down{1.3} & \down{6.2} & \down{0.2} & \down{3.7} \\
& S250 & 2.49 & 14.79 & 8.94 & 9.01 & 9.03 & 65.0 & 41.6 & 73.7 & 70.6 & 77.9 & 93.1 & 65.4 & 42.0 & 66.2
\\[-1pt]
& {\footnotesize $\Delta$P200-R50} & \up{0.05} & \up{0.55} & \up{0.36} & \up{0.39} & \up{0.43} & \up{0.6} & \up{0.9} & \up{2.2} & \up{2.6} & \up{1.1} & \up{1.4} & \up{1.3} & \up{4.0} & \up{1.8} \\
\noalign{\vspace{-3pt}}\midrule
\multirow{5}{*}{\makecell[l]{Minitron\\width}}
& P200-R50 & 2.47 & 14.60 & 8.93 & 9.01 & 8.92 & 65.4 & 41.8 & 73.4 & 70.5 & 77.4 & 93.8 & 67.4 & 40.4 & 66.3 \\[3pt]
& S50 & 2.70 & 17.47 & 10.99 & 11.17 & 10.65 & 61.3 & 33.6 & 65.1 & 60.3 & 74.7 & 89.6 & 60.8 & 33.8 & 59.9
\\[-1pt]
& {\footnotesize $\Delta$P200-R50} & \down{0.23} & \down{2.87} & \down{2.05} & \down{2.16} & \down{1.73} & \down{4.1} & \down{8.2} & \down{8.2} & \down{10.2} & \down{2.7} & \down{4.2} & \down{6.6} & \down{6.6} & \down{6.4} \\
& S250 & 2.46 & 14.37 & 8.62 & 8.69 & 8.75 & 65.8 & 46.2 & 75.6 & 72.2 & 77.9 & 91.2 & 68.4 & 42.6 & 67.5
\\[-1pt]
& {\footnotesize $\Delta$P200-R50} & \same{0.00} & \up{0.23} & \up{0.32} & \up{0.33} & \up{0.17} & \up{0.4} & \up{4.4} & \up{2.2} & \up{1.7} & \up{0.5} & \down{2.6} & \up{1.0} & \up{2.2} & \up{1.2} \\
\noalign{\vspace{-3pt}}\midrule
\multirow{5}{*}{\makecell[l]{FLAP}}
& P200-R50 & 2.47 & \best{13.64} & \best{7.90} & \best{7.97} & 8.55 & 65.9 & 43.5 & \best{75.8} & 72.8 & 76.9 & 93.9 & \best{75.5} & 27.6 & 66.5 \\[3pt]
& S50 & 2.71 & 17.57 & 11.16 & 11.34 & 10.77 & 58.3 & 34.9 & 67.2 & 59.8 & 73.8 & 83.3 & 58.9 & 25.4 & 57.7
\\[-1pt]
& {\footnotesize $\Delta$P200-R50} & \down{0.24} & \down{3.93} & \down{3.26} & \down{3.37} & \down{2.22} & \down{7.6} & \down{8.6} & \down{8.6} & \down{13.0} & \down{3.1} & \down{10.6} & \down{16.6} & \down{2.2} & \down{8.8} \\
& S250 & 2.51 & 14.57 & 8.76 & 8.82 & 8.92 & 66.1 & 43.2 & 74.8 & 71.7 & 77.4 & 91.1 & 66.8 & 29.8 & 65.1
\\[-1pt]
& {\footnotesize $\Delta$P200-R50} & \down{0.04} & \down{0.93} & \down{0.86} & \down{0.85} & \down{0.37} & \up{0.2} & \down{0.3} & \down{1.0} & \down{1.1} & \up{0.5} & \down{2.8} & \down{8.7} & \up{2.2} & \down{1.4} \\
\noalign{\vspace{-3pt}}\midrule
\multirow{5}{*}{\makecell[l]{Sheared\\LLaMA}}
& P200-R50 & 2.50 & 14.35 & 8.55 & 8.63 & 8.90 & 65.8 & 42.8 & 74.7 & 71.2 & 76.7 & 93.6 & 68.0 & 34.2 & 65.9 \\[3pt]
& S50 & 2.67 & 17.51 & 11.10 & 11.28 & 10.69 & 60.0 & 34.2 & 66.1 & 59.9 & 74.6 & 89.0 & 59.6 & 23.8 & 58.4
\\[-1pt]
& {\footnotesize $\Delta$P200-R50} & \down{0.17} & \down{3.16} & \down{2.55} & \down{2.65} & \down{1.79} & \down{5.8} & \down{8.6} & \down{8.6} & \down{11.3} & \down{2.1} & \down{4.6} & \down{8.4} & \down{10.4} & \down{7.5} \\
& S250 & 2.47 & 14.57 & 8.77 & 8.85 & 8.89 & 64.8 & 44.2 & 74.7 & 71.6 & 77.9 & 91.9 & 67.1 & 30.4 & 65.3
\\[-1pt]
& {\footnotesize $\Delta$P200-R50} & \up{0.03} & \down{0.22} & \down{0.22} & \down{0.22} & \up{0.01} & \down{1.0} & \up{1.4} & \same{0.0} & \up{0.4} & \up{1.2} & \down{1.7} & \down{0.9} & \down{3.8} & \down{0.6} \\
\noalign{\vspace{-3pt}}\midrule
\multirow{5}{*}{\makecell[l]{Wanda\\2:4}}
& P200-R50 & 2.49 & 14.71 & 9.07 & 9.15 & 8.96 & 66.1 & 43.6 & 74.5 & 70.5 & 78.8 & 93.5 & 66.7 & 40.2 & 66.7 \\[3pt]
& S50 & 2.63 & 16.68 & 10.40 & 10.55 & 10.23 & 59.2 & 36.8 & 70.3 & 62.8 & 75.4 & 90.4 & 60.4 & 37.0 & 61.5
\\[-1pt]
& {\footnotesize $\Delta$P200-R50} & \down{0.14} & \down{1.97} & \down{1.33} & \down{1.40} & \down{1.27} & \down{6.9} & \down{6.8} & \down{4.2} & \down{7.7} & \down{3.4} & \down{3.1} & \down{6.2} & \down{3.2} & \down{5.2} \\
& S250 & 2.45 & 14.33 & 8.54 & 8.60 & 8.74 & 64.2 & 42.4 & 74.4 & 70.3 & 77.1 & 92.3 & 68.5 & 40.4 & 66.2
\\[-1pt]
& {\footnotesize $\Delta$P200-R50} & \up{0.04} & \up{0.39} & \up{0.53} & \up{0.54} & \up{0.22} & \down{1.8} & \down{1.2} & \down{0.1} & \down{0.2} & \down{1.7} & \down{1.2} & \up{1.8} & \up{0.2} & \down{0.5} \\
\noalign{\vspace{-3pt}}\midrule
\multirow{5}{*}{\makecell[l]{Wanda-U}}
& P200-R50 & \best{2.43} & 13.94 & 8.31 & 8.37 & \best{8.47} & 67.7 & \best{46.7} & \best{75.8} & \best{73.1} & \best{79.0} & \best{94.9} & 66.7 & 40.8 & \best{68.1} \\[3pt]
& S50 & 2.63 & 16.72 & 10.51 & 10.66 & 10.26 & 60.1 & 36.3 & 69.5 & 62.9 & 75.9 & 90.5 & 60.0 & 37.2 & 61.6
\\[-1pt]
& {\footnotesize $\Delta$P200-R50}  & \down{0.21} & \down{2.78} & \down{2.21} & \down{2.30} & \down{1.80} & \down{7.6} & \down{10.3} & \down{6.3} & \down{10.2} & \down{3.1} & \down{4.4} & \down{6.7} & \down{3.6} & \down{6.5} \\
& S250 & 2.45 & 14.29 & 8.51 & 8.57 & 8.73 & 64.7 & 43.6 & 75.5 & 72.1 & 78.8 & 94.2 & 67.1 & 41.6 & 67.2
\\[-1pt]
& {\footnotesize $\Delta$P200-R50} & \down{0.02} & \down{0.35} & \down{0.20} & \down{0.20} & \down{0.26} & \down{3.0} & \down{3.1} & \down{0.3} & \down{1.0} & \down{0.2} & \down{0.7} & \up{0.4} & \up{0.8} & \down{0.9} \\
\noalign{\vspace{-3pt}}\midrule
\multirow{5}{*}{\makecell[l]{SparseGPT\\2:4}}
& P200-R50 & 2.48 & 14.60 & 8.83 & 8.90 & 8.90 & 66.5 & 42.3 & 73.2 & 70.6 & 78.1 & 93.7 & 67.7 & \best{41.6} & 66.7 \\[3pt]
& S50 & 2.63 & 16.72 & 10.40 & 10.54 & 10.25 & 62.2 & 40.4 & 70.2 & 62.7 & 75.4 & 90.6 & 58.6 & 36.6 & 62.1
\\[-1pt]
& {\footnotesize $\Delta$P200-R50} & \down{0.15} & \down{2.12} & \down{1.57} & \down{1.64} & \down{1.35} & \down{4.3} & \down{2.0} & \down{2.9} & \down{8.0} & \down{2.7} & \down{3.1} & \down{9.1} & \down{5.0} & \down{4.6} \\
& S250 & 2.45 & 14.32 & 8.59 & 8.65 & 8.72 & 67.1 & 42.0 & 73.5 & 71.1 & 78.9 & 95.0 & 68.8 & 40.0 & 67.1
\\[-1pt]
& {\footnotesize $\Delta$P200-R50} & \up{0.03} & \up{0.28} & \up{0.24} & \up{0.25} & \up{0.18} & \up{0.6} & \down{0.3} & \up{0.3} & \up{0.5} & \up{0.8} & \up{1.3} & \up{1.1} & \down{1.6} & \up{0.4} \\
\noalign{\vspace{-3pt}}\midrule
\multirow{5}{*}{\makecell[l]{SparseGPT-U}}
& P200-R50 & 2.44 & 14.06 & 8.39 & 8.44 & 8.56 & \best{68.3} & 45.0 & 75.3 & 72.3 & 78.9 & 94.1 & 65.9 & 40.0 & 67.5 \\[3pt]
& S50 & 2.63 & 16.65 & 10.31 & 10.47 & 10.22 & 60.4 & 36.9 & 69.3 & 62.5 & 75.0 & 90.6 & 62.5 & 38.2 & 61.9
\\[-1pt]
& {\footnotesize $\Delta$P200-R50}  & \down{0.19} & \down{2.59} & \down{1.93} & \down{2.02} & \down{1.66} & \down{7.9} & \down{8.1} & \down{6.0} & \down{9.8} & \down{4.0} & \down{3.5} & \down{3.4} & \down{1.8} & \down{5.5} \\
& S250 & 2.45 & 14.31 & 8.53 & 8.59 & 8.74 & 67.0 & 43.2 & 73.7 & 71.0 & 78.2 & 93.1 & 62.4 & 41.6 & 66.3
\\[-1pt]
& {\footnotesize $\Delta$P200-R50} & \down{0.01} & \down{0.26} & \down{0.15} & \down{0.15} & \down{0.18} & \down{1.3} & \down{1.8} & \down{1.6} & \down{1.3} & \down{0.8} & \down{1.0} & \down{3.5} & \up{1.6} & \down{1.2} \\
\noalign{\vspace{-3pt}}\bottomrule
\end{tabular}
}
\end{adjustbox}
\caption{\textbf{Pruning-retraining versus training from scratch at 50\% pruning ratio.} \textit{P200-R50:} pretrain 200B tokens, prune, retrain 50B tokens. \textit{S50:} train from scratch for 50B tokens (equal training token budget). \textit{S250:} train from scratch for 250B tokens (equal total token budget). $\Delta$P200-R50: difference relative to P200-R50. Red: P200-R50 is better; blue: P200-R50 is worse; grey: within evaluation standard deviation (see Appendix~\ref{appendix:evaluation}).}
\label{table:pruning-vs-scratch-50pct}
\end{table*}

\paragraph{Per-benchmark observations.} FLAP exhibits the strongest BoolQ initialization signal in the table: $\Delta\text{BoolQ}\!=\!+16.6$ vs.\ S50, and it is the only benchmark/method combination where the residual advantage at S250 exceeds 8 points ($+8.7$). FLAP also holds the best-cell perplexity numbers overall (C4 13.64, WT 7.90, WT-2 7.97), surpassing every sparse method despite being structured --- a worth-noting caveat to the headline ``sparse $>$ structured at full data'' framing. OBQA splits the sparse and structured methods at S250: Sheared LLaMA retains $+3.8$ (largest residual structured advantage) while Wanda-U and SparseGPT-2:4 actually trail S250 on OBQA. Finally, the sparse-method advantage at full data is benchmark-dependent rather than uniform: Wanda-2:4, Wanda-U, and SparseGPT-U all post slightly negative average deltas at S250 (down to $-1.2$), even though individual benchmarks (BoolQ, WT-2 perplexity) still favor pruning --- readers should treat the sparse advantage under an equal total token budget as a per-benchmark rather than a wholesale claim.

\end{document}